\pdfoutput=1
\documentclass[11pt,a4paper]{article}
\usepackage[]{geometry}

\usepackage{graphicx}
\usepackage{amsmath}
\usepackage{cite}
\usepackage{xcolor}
\usepackage[font=small,labelfont=bf]{caption}

\usepackage{pdflscape}  
\usepackage{tabularx}
\usepackage{booktabs}
\usepackage{arydshln}   
\usepackage{makecell, multirow}
\usepackage{adjustbox}
\usepackage{listings}   
\usepackage{xurl}       
\providecommand{\doi}[1]{\href{https://doi.org/\detokenize{#1}}{\url{https://doi.org/#1}}}

\usepackage[hidelinks]{hyperref} 

\usepackage[T1]{fontenc}

\newcommand{\linkref}[3]{\hyperref[#1]{#2~\ref*{#1}#3}}

\usepackage{xspace}  

\newcommand{\supWikipedia}{\linkref{appendix_wikipedia}{Appendix}{}\xspace}
\newcommand{\supOpenalex}{\linkref{appendix_openalex}{Appendix}{}\xspace}
\newcommand{\supGraph}{\linkref{appendix_datasets}{Appendix}{}\xspace}
\newcommand{\supTFIDF}{\linkref{appendix_tfidf}{Appendix}{}\xspace}
\newcommand{\supEmbedding}{\linkref{appendix_embedding_models}{Appendix}{}\xspace}
\newcommand{\supLLM}{\linkref{appendix_llm_models}{Appendix}{}\xspace}
\newcommand{\supLLMExplain}{\linkref{appendix_llm_wiki}{Appendix}{}\xspace}
\newcommand{\supPrompt}{\linkref{appendix_llm_prompt}{Appendix}{}\xspace}
\newcommand{\supPerformance}{\linkref{appendix_model_performance}{Appendix}{}\xspace}
\newcommand{\supDistribution}{\linkref{appendix_distribution_fitting}{Appendix}{}\xspace}
\newcommand{\supLinkTitles}{\linkref{appendix_title_performance}{Appendix}{}\xspace}
\newcommand{\supFailure}{\linkref{appendix_failure_reasons}{Appendix}{}\xspace}
\newcommand{\supCorpus}{\linkref{appendix_tfidf_corpus}{Appendix}{}\xspace}
\newcommand{\supRoutes}{\linkref{appendix_routes}{Appendix}{}\xspace}
\newcommand{\supMultilingual}{\linkref{appendix_oa_multilingual}{Appendix}{}\xspace}
\newcommand{\supPostTraining}{\linkref{appendix_post_2024}{Appendix}{}\xspace}

\title{Can Language Models Play the Wikipedia Game?}

\author{Alexander Hickey, Joanna Knight, and James Bishop\\[0.75ex]
\normalsize The Alan Turing Institute, 96 Euston Rd., London NW1 2DB, UK\\
\normalsize\texttt{ahickey@turing.ac.uk}}
\date{}

\begin{document}

\maketitle

\begin{abstract}
    \typeout{~~ Abstract ~~}

Language models increasingly support or automate information retrieval from document collections too large to index.
This task is difficult when it requires specialist linguistic or scientific knowledge.
We used the Wikipedia Game to test language models on an analogous information retrieval task.
In the game, players must reach a target article using only article links.
We compared sentence embedding models and LLMs across five Wikipedia languages and five academic-literature datasets.
We construct very large local graphs of Wikipedia pages and of paper citations using OpenAlex and provide code to build and navigate them.
This allows the study of language models' navigation capabilities without relying on third-party APIs.
We compared the models with human and TF-IDF benchmarks.
The Wikipedia Game format enables robust comparisons across techniques, languages, and domains.
No model significantly outperformed humans, although the best models (GPT-o3 mini, GPT-4o, QwQ:32B, and DeepSeek-R1:70B) achieved comparable success.
Performance and rankings varied across languages and topics, with implications for practitioners utilising these techniques.
\end{abstract}

\section{Introduction}
\typeout{~~ Introduction ~~}

This project investigates the capabilities and limitations of language models when automatically navigating the content of unstructured document stores.
Language models are often used to automate time-consuming tasks, and many of these, such as information retrieval, text classification, and question answering, require the models to connect concepts semantically~\cite{kimBridgingGapIncorporating2017,chandrasekaranEvolutionSemanticSimilarity2021,smeatonUsingNLPNLP1999,kimConvolutionalNeuralNetworks2014,allamQuestionAnsweringSystems2012}.
We use the Wikipedia Game as an analogous task and present a comparison of human performance alongside an evaluation of the capabilities of various models across different topics and languages.
This work constructed multiple vast and navigable graph datasets and the code to do so is available at: \url{https://github.com/alan-turing-institute/ARC-Wiki-Game}

Wikipedia is a free online encyclopedia that contains millions of hyperlinked articles created by over a quarter of a million editors and is available in over 300 languages.
In the Wikipedia Game, participants start with two random articles and are expected to navigate between them by clicking only on links within each article.
Succeeding in the Wikipedia Game requires players to continuously evaluate the relatedness between the target article and the linked articles available.
While sentence embedding models are naturally built to assess the similarity between segments of text, large language models (LLMs) might also possess the ability to simulate strategies, which could provide a competitive advantage~\cite{huangUnderstandingPlanningLLM2024}.
Throughout this report, LLM refers to a generative prompt-based model like GPT-4o or Llama.

Previous literature has discussed the Wikipedia Game, and recent natural language processing (NLP) research has started to evaluate LLMs on the game.
So far, all this work has assessed the dataset via an API \cite{margiottaEvaluatingLargeLanguage2025} or has used a small subset of the full graph \cite{ebrahimiTextualUnderstandingBoost2026,nogueiraEndtoEndGoalDrivenWeb2016,westAutomaticHumanNavigation2012,ziomekLLMWikiRaceBenchmarkHow2026}; this is also true of demos produced by HuggingFace and TypeSafe AI.
This project explores the ability of language models to play these games on Wikipedia articles in a variety of languages and on a dataset of academic works, OpenAlex\footnote{OpenAlex - \url{https://openalex.org/}}.
By building these graphs to facilitate utilising these datasets, we can see how varieties of language models perform across languages and subject areas, generating insights relevant to real-world applications \cite{chakrabartiAcceleratedFocusedCrawling2002}.
These findings can help inform individuals looking to incorporate these models into their data pipelines.
\subsection{Wikipedia as a Graph}

English Wikipedia has over six million articles, which contain nearly 300 million links between them, making it the largest reference work in history.
This encyclopedia can be conceptualised as a graph comprising connected nodes (the articles) and edges (the hyperlinks between the articles) \cite{consonniWikiLinkGraphsCompleteLongitudinal2019}.
Doing so has enabled other researchers to construct a knowledge graph~\cite{wangWikiGraphsWikipediaText2021} or identify useful missing links~\cite{shaptalaUsingGraphEmbeddings2019}.
As Wikipedia is both human-written text and organised within a structured format, it has been harnessed for topic classification~\cite{johnsonLanguageagnosticTopicClassification2021}, improving the performance of foundation models~\cite{yasunagaLinkBERTPretrainingLanguage2022}, and providing reliable ground truth data for named entity recognition~\cite{balasuriyaNamedEntityRecognition2009}.

\subsection{Wikipedia as a Testing Ground}

This work leverages the size and structure of Wikipedia and similar datasets to provide a robust challenge for language models to utilise the semantic information they contain.
Previous research has capitalised on these properties of Wikipedia to assess word embeddings and graph search algorithms and construct benchmarks~\cite{giesenLeveragingWikipediaGraph2022,hewlettWikiReadingNovelLargescale2017,barronAIWikipediaGame2015}.
Other studies have leveraged human performance in the Wikipedia game as a source of semantic information \cite{westWikispeediaOnlineGame2009,westExtractingSemanticInformation2010}.
This work builds on these previous studies by assessing sentence embeddings and LLMs, making it possible to compare NLP methods and human abilities.

Beyond language models, Wikipedia is an excellent training dataset for graph neural networks~\cite{sanchez-lengelingGentleIntroductionGraph2021,mernyeiWikiCSWikipediaBasedBenchmark2020}.
Additionally, these same techniques can be applied to citation networks in academic literature to more accurately assign papers to an academic topic~\cite{lyArticleClassificationGraph2024}.

\subsection{Generative and Embedding Models}

One may naively consider the Wikipedia Game to be a sequence of judging how related each potential next article is to the target article.
Sentence embedding models are particularly well-suited for this task~\cite{caoRecentAdvancesText2024} as these models aim to produce a vector that numerically represents the meaning conveyed by a given sentence~\cite{reimersSentenceBERTSentenceEmbeddings2019}.
Embedding models can be optimised to connect related concepts or to focus on distinguishing topics or sentiments~\cite{asudaniImpactWordEmbedding2023}, work in multiple languages~\cite{reimersMakingMonolingualSentence2020} and address specialised domains~\cite{pankajAugmentedBioSBERTImproving2022}.

When people play the Wikipedia Game, they do not just jump incrementally to the most related article but are instead planning a route ahead.
This kind of planning and strategy cannot be reproduced by an embedding model that relies solely on a greedy iterative approach that always seeks the next `closest' article.
Reinforcement Learning (RL) can be used to incorporate strategic considerations into deep learning techniques, and there are numerous examples of combining RL with NLP techniques (e.g. playing text adventure games~\cite{osborneSurveyTextGames2022,narasimhanLanguageUnderstandingTextbased2015}).

There is some evidence that LLMs can plan and show strategic considerations ~\cite{huangUnderstandingPlanningLLM2024,zhangLLMMastermindSurvey2024}, which could give an advantage over embedding models.
However, there are many examples where their strategic considerations fall short of what might be expected based on their language comprehension ability~\cite{valmeekamPlanningAbilitiesLarge2023,nezhurinaAliceWonderlandSimple2024}.
This issue may arise because the majority of LLM training is based on next-word prediction; it is theorised that the model's reasoning processes quickly deteriorate once it is outside examples provided in the training data~\cite{mondorfAccuracyEvaluatingReasoning2024,xieMemorizationLargeLanguage2024}.
Comparing LLMs' performance at this task to sentence embedding models, which can only provide semantic comparisons of sentences, might help illuminate their strategic ability.

\begin{figure}[t!]
    \centering
    \includegraphics[width=\linewidth]{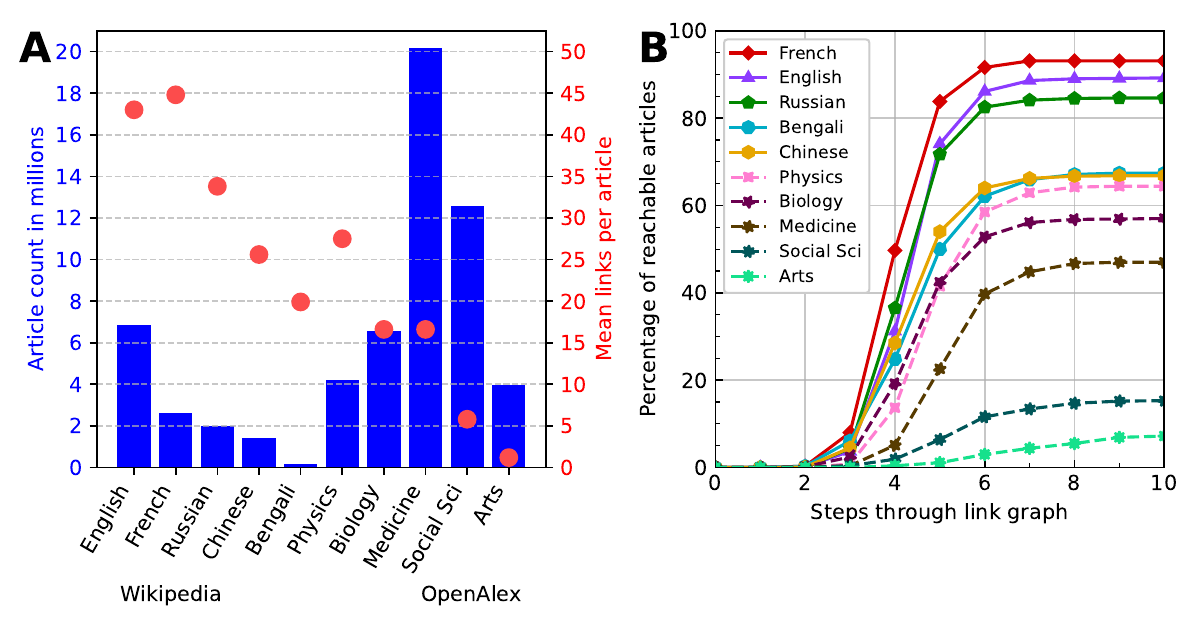}
    \caption{\protect\input{figures/fig1_caption}}
    \label{fig1:graph}
\end{figure}

\section{Datasets}
\typeout{~~ Datasets ~~}

Most of our results are presented for English Wikipedia; however, we tested the models on a collection of different graph datasets so that the performance across languages and technical subjects can be explored.
Five datasets in different languages were obtained from Wikipedia, while the subject-specific datasets were compiled from a collection of abstracts in the scientific literature.
Statistics comprising these datasets can be found in \supGraph.

\subsection{Wikipedia in Many Languages}

English Wikipedia contains over six million articles and four and a half billion words.\footnote{Size of Wikipedia - \url{https://en.wikipedia.org/wiki/Wikipedia:Size_of_Wikipedia}}
These articles cross-reference each other with links, which produces a navigable graph across a large fraction of the articles.

We used Wikipedia in five different languages which are shown in \linkref{fig1:graph}{Figure}{A} with more details in \supWikipedia.
Four of these are considered high-resource languages (English, French, Chinese, and Russian), whereas Bengali is considered a lower-resource language \cite{mahfuzTooLateTrain2024}.
A directed graph is created for each language using the Wikipedia datasets, with the nodes representing articles and the edges representing links between articles.
The title and a summary containing the start of the article are stored with each node.
Note that the graph is not symmetric: if Article A has a link to Article B, this does not guarantee that Article B contains a link to Article A.

\subsection{OpenAlex}

OpenAlex is a dataset of metadata (including the title, authors, abstract, and references) about academic works.
The OpenAlex data have been used to create graphs with the references and citations as the links between the articles, and the abstracts were used like the summaries.
To create datasets more similar to Wikipedia, this project limited the type of work to `articles' (other types of work that were excluded include `book chapters', `dissertations' and `letters').
Unlike the Wikipedia dataset, the OpenAlex datasets are symmetric graphs because if Article A references Article B, then Article B will be cited by Article A.

The OpenAlex dataset was categorised based on each paper's primary academic field to create multiple graphs (additional details in \supOpenalex).
Using the primary topic categorisation, five subjects were chosen (see \linkref{fig1:graph}{Figure}{A}).
The connectivity of the graphs reflects the referencing patterns within each academic field, leading to significant variations in the graphs' navigability.
\linkref{fig1:graph}{Figure}{B} illustrates the proportion of the graph that is reachable after a given number of steps.
To account for these differences, only navigable pairs of articles were used to test the language models.

\section{Experimental Setup}
\typeout{~~ Exprimental Setup ~~}

For each of the datasets listed in \linkref{fig1:graph}{Figure}{A}, a set of 10,000 pairs of articles were randomly generated, with each pair representing one `game'.
One article in each pair was labelled the source article and the other the target article; the pair was discarded if it was not possible to navigate between them.

The goal of the game is to navigate to the target article using only the links in the graph.
In the two formulations of the game, the agent player may only see the title of each link or the title and the summary text of each possible link (resembling the mouse-over preview in Wikipedia).
In both formulations, the agent is allowed to see the summary of the target article.
Only the choices on the `current' article can be viewed before one is selected: the agent is not allowed to look further ahead or to work backwards from the target article.

The same set of games was used for each agent, but resource constraints meant that not all agents could play all 10,000 games.
For the experiments using OpenAlex, the target articles were filtered on the language field of `English'; however, the relaxation of this constraint is explored in \supMultilingual.

When the agents played the game, there were a few ways that it could end, the simplest being that the target article was reached successfully.
However, the agent can fail to reach the target article in multiple ways, the most common being not reaching the target within 100 steps.
It was also possible for the model to reach a dead end, for the LLM-based agents to repeatedly generate an article title that does not exist or reject the request, and for reasoning models to fail to provide an answer.
More details on this are in \supFailure.

\section{Agents}
\typeout{~~ Agents ~~}

This work considers two types of language models: sentence embedding models and LLMs.
We compare their ability to play the Wikipedia Game to a statistical approach and human attempts.
The setup for all these methods is optimised to compare how these techniques might be used in a real-world setting.

\subsection{Sentence Embeddings}

Sentence embedding models map text to numerical vectors within a high-dimensional space where semantically similar sentences are positioned closely together~\cite{kashyapComprehensiveSurveySentence2024}.
We designed an agent that used a pre-trained sentence embedding model to select the link at each stage of the game with the `closest' summary sentence or title to the target summary sentence.
The `closest' link was determined by the minimum cosine distance.
As with all the agents, additional logic prevents them from returning to an article they have already visited to avoid becoming trapped in loops. 

It is possible to play the game either by comparing the summaries or the titles of the available articles.
Both of these versions of the game were considered, and the results here focus on using the summary embeddings.
The impact on performance when only using the titles can be found in \supLinkTitles.

To aid readability, we will focus on the performance of five of the thirteen embedding models studied in this work.
E5-Small is a multilingual embedding model produced by Microsoft~\cite{wangMultilingualE5Text2024}.
AllMPNetBase is recommended by the Sentence Transformers documentation and is a comparable size to E5-Small, but is only trained on English language data~\cite{reimersSentenceBERTSentenceEmbeddings2019}.
MiniLM-L6 was one of the smallest models considered with only 22.7~million parameters~\cite{wangMiniLMDeepSelfAttention2020,reimersMakingMonolingualSentence2020}.
Roberta makes use of the well-known BERT architecture and is fine-tuned to produce an embedding model~\cite{reimersMakingMonolingualSentence2020,liuRoBERTaRobustlyOptimized2019a}.
Glove works by averaging the individual word embedding, which makes it much faster in inference than similarly sized models~\cite{penningtonGloVeGlobalVectors2014,reimersSentenceBERTSentenceEmbeddings2019}.
The other models are listed in \supEmbedding.

\subsection{Prompting LLMs}
\label{subsection_prompt_agent}

LLMs used here generate human-like text by predicting each word incrementally.
By crafting detailed and specific starting text, known as the prompt, LLMs can be directed to perform precise tasks such as code synthesis, text labelling, and machine translation~\cite{jimenezEvaluationLLMCode2024,myrzakhanOpenLLMLeaderboardMultichoiceOpenstyle2024,topsakalBenchmarkingLargeLanguage2024}.
Including all the necessary information in the prompt is crucial for maximising the capabilities of these models.

This work used thirteen different LLMs (listed in \supLLM), including the full suite of Llama3 models, which range from 1 to 405~billion parameters \cite{grattafioriLlama3Herd2024,llamateamaiatmetaLlama32Revolutionizing2024}.
A Gemma2 model was also used, which has 27~billion parameters~\cite{gemmateamGemma2Improving2024}.
OpenAI's GPT-4o was the largest LLM used in this work with its smaller version, GPT-4o mini.
However, due to the cost of running GPT-4o, the number of games played using this agent was limited to 100 and restricted to only a few datasets.
There were also 5 "reasoning (R)" LLMs evaluated, which included DeepSeek(DS) R1~\cite{guoDeepSeekR1IncentivizesReasoning2025} and Qwen with Questions (QwQ).

\supLLMExplain shows that the larger LLMs are able to explain how to play the Wikipedia game. 
The agent prompts an LLM by informing it that it is participating in the Wikipedia Game, providing a summary of the target article, and asking which article to choose from a list of titles.
This requires a significant amount of information to be provided to the LLM, which can negatively affect its performance~\cite{levySameTaskMore2024}.
Two key adjustments were implemented to address this issue.
Firstly, a maximum limit of 500 links for GPT and 200 links for the other models from the current article was established.
In preliminary experiments, GPT-4o was able to cope with a longer prompt and still succeeded, but the success rate of all LLMs improved when the number of titles was limited.
Secondly, the prompt included only the titles of the linked articles rather than the article summaries.
These adaptions greatly improved the reliability of the prompting method.

As an LLM can, in principle, output any text, a title that does not match any in the list provided may be generated.
In this case, the LLM is prompted again up to four more times with an extra paragraph emphasising the task.
The step counts shown in the results record the number of articles visited, not the number of times the LLM is queried.
For experimental consistency, the same prompt was used for all non-reasoning LLMs, and the article limit was set high enough to only affect a small minority of articles.
Further details can be found in \supPrompt.

\subsection{Reasoning LLMs}

This work also considered evaluating an LLM in reasoning mode where the LLM can generate a `thinking' block of text before providing the answer.
We evaluated eight of these LLMs in this way.
For GPT-o3 mini, this was done by using the same prompt as with the GPT-4o model, but allowing reasoning through the API.
The open-source models were given chain-of-thought prompting and could provide an answer (contained in \texttt{<answer>} tags) in the output text.
The DeepSeek (DS) and QwQ models are trained for this behaviour, while the Llama and Gemma models were not.
To distinguish which setting the models were evaluated in, an (R) is placed after the name of the models evaluated in the reasoning format.
Other than the requested format, the experiment was the same for the reasoning and non-reasoning LLMs.

\subsection{TF-IDF}

A non-deep-learning benchmark was created using term frequency-inverse document frequency (TF-IDF) - a statistical approach that assesses the importance of a word to a document based on its frequency in the corpus.
For each dataset, a corpus was created using a sample of 100,000 articles from that dataset.
This does mean that some of the corpora may be more representative of some datasets than others; for example, 100,000 articles is a greater proportion of the Bengali Wikipedia dataset than the English.
The articles used in the sample are not held separately and may appear in the game.
Details on how the TF-IDF values are calculated can be found in \supTFIDF.

\subsection{Humans}

Nine human players attempted a sample of games for the English Wikipedia dataset to act as a benchmark.
These games were not played under strict experimental conditions, and the players were requested to self-report their success.
We used the website WikiSpeedrun\footnote{WikiSpeedrun - \url{https://wikispeedrun.org/}} to provide a timer and count of steps taken. 
To increase the number of games played, a time limit of 10 minutes was used instead of a step limit, with players aware of this limit.
In total, 109 games were played, and an overlap of games was avoided between the players.

\begin{figure}[t!]
    \centering
    \includegraphics[width=\linewidth]{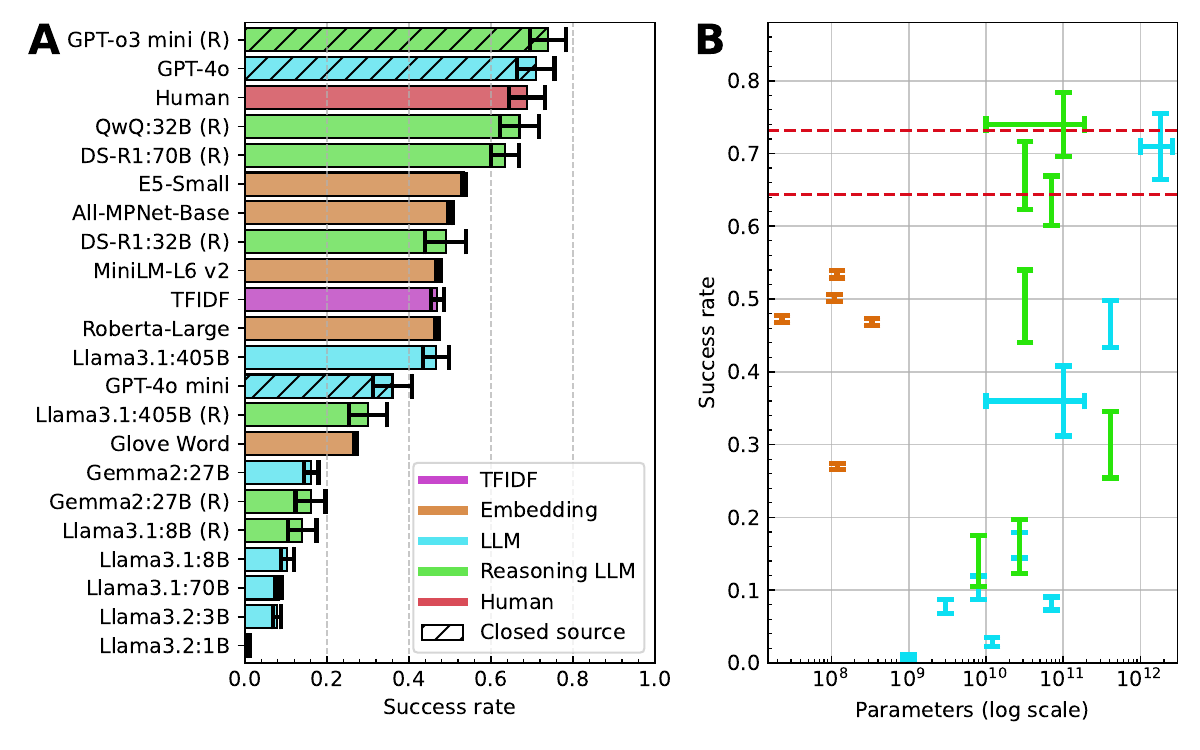}
    \caption{\protect\input{figures/fig2_caption}}
    \label{fig2:performance}
\end{figure}

\section{Model Performance}
\label{section_model_performance}
\typeout{~~ Model Performance ~~}

English Wikipedia is used in this section to assess all the approaches; a table of all the results can be found in \supPerformance.

\subsection{Success Rate}

\linkref{fig2:performance}{Figure}{A} shows the success rate of a selection of the models on the English Wikipedia graph.
A game was considered a success if the model reached the target article from the start article in 100 steps.
\linkref{fig2:performance}{Figure}{A} shows that GPT-o3 mini (R) had the highest success rate, comparable to that for humans.

Many sentence embedding models were able to perform similarly to the largest of the open non-reasoning LLMs, as shown in \linkref{fig2:performance}{Figure}{B}.
The optimum formulation was used for both model types, so the embedding models used summaries of the linked articles, whereas the LLMs were only shown the title.
\supLinkTitles shows there was a measurable but not catastrophic drop in the success rate when the embedding models were only shown the title of the articles.

Unlike the embedding models, the performance of the LLMs severely deteriorated with decreasing model size, which may be related to the model's ability to stay on task reliably.
This is seen in the much higher tendency for the smaller LLMs to hallucinate.
Reasoning models did manage to provide more accurate answers for the same-sized LLM, but the smallest reasoning model, DS-R1:8b, did not complete the task.
\supFailure shows the reasons for failure by the different models.

TF-IDF was also effective at the task.
The corpus for the TF-IDF model was generated with a sample from the article summaries used to play the game, and it is likely that this technique would not be as performant if this construction was not possible.
\supCorpus shows the success rate for TF-IDF models on the OpenAlex datasets when English Wikipedia was used as the corpus, as evidence for this.

\subsection{Efficiency}

\begin{figure}[t!]
    \centering
    \includegraphics[width=\linewidth]{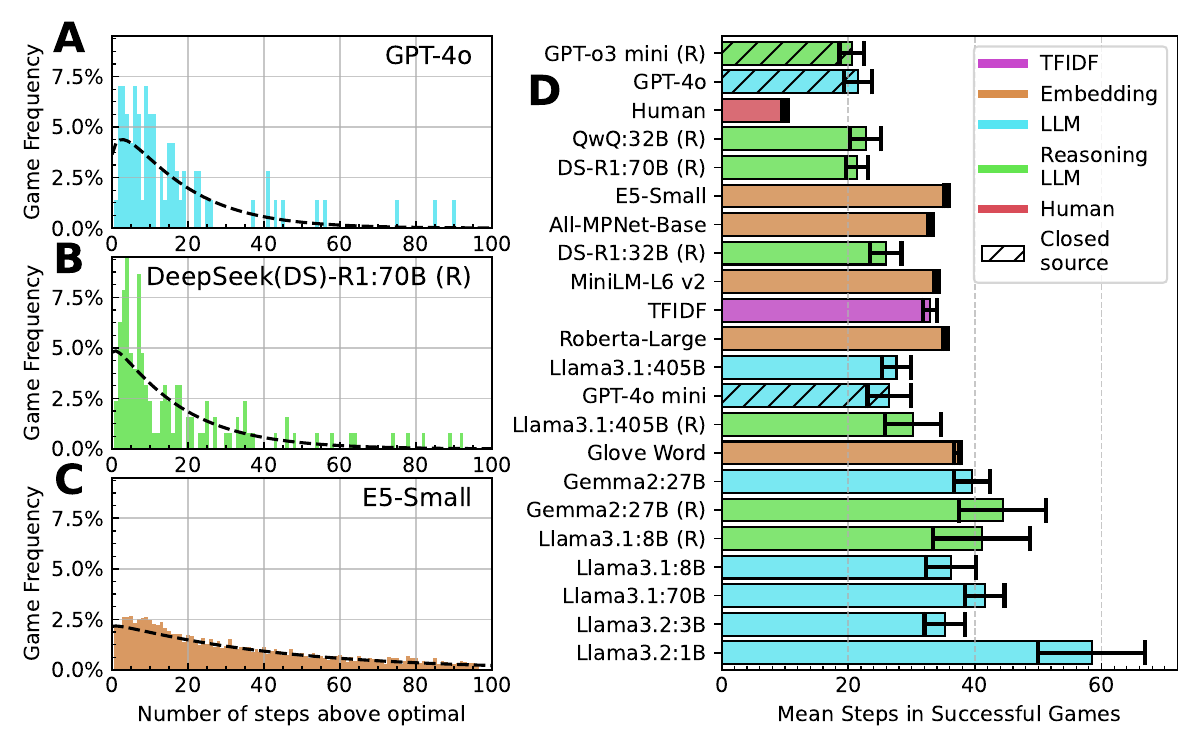}
    \caption{\protect\input{figures/fig3_caption}}
    \label{fig3:step_count}
\end{figure}

\linkref{fig3:step_count}{Figure}{} examines the efficiency of the various agents for the games where they succeeded, as measured by the number of steps taken to reach the target article.
\hyperref[fig3:step_count]{Figures \ref*{fig3:step_count}A, \ref*{fig3:step_count}B, and \ref*{fig3:step_count}C} show histograms of the number of extra steps taken above the optimal route for GPT-4o, DS-R1:70B (R) and E5-Small, which seems to follow a negative binomial distribution.
The figures show that GPT-4o can succeed in fewer steps for a greater proportion of games than E5-Small.
This reflects differences in how the types of models played the game.
An example of the routes taken by E5-Small and GPT-4o on the same game illustrates this in \supRoutes.

\linkref{fig3:step_count}{Figure}{D} shows a clear distinction between human performance and that of GPT and DS-R1; even though both approaches achieved a comparable success rate, humans were able to succeed with fewer steps.
As the LLM agents allowed five retries in the event of hallucinations, the LLM models can be considered even less efficient if the number of queries is counted rather than the number of steps.
The LLMs of Llama:405B and GPT-4o mini can be more efficient than the sentence embedding models.
Trained reasoning models were also comparable in efficiency to the non-reasoning LLMs of similar performance.
This may be indicative of strategies for playing the Wikipedia Game being taken into account by LLMs over the embedding models.
The language models' step distributions did not differ measurably from a geometric distribution (i.e. a negative binomial with $n \approx 1$), but the human games do, even when games more than 10 steps above optimal are ignored (\supDistribution).

\section{Languages and Topics}
\label{section_lang_topics}
\typeout{~~ Languages and Topics ~~}

\linkref{fig4:topics}{Figure}{} shows the success rate for a selection of the models across the other datasets of Wikipedia languages and OpenAlex topics - the success rates for all the models which were evaluated can be found in \supPerformance. 

\subsection{TF-IDF and Embedding Models}

\linkref{fig4:topics}{Figure}{} compares two sentence embedding models with the TF-IDF approach. The E5-Small embedding model achieved the highest success rates (within a margin of error) across all five languages of the Wikipedia datasets.
The TF-IDF approach is sometimes competitive with the embedding models, and provides a good statistical comparison across the datasets.
However, \linkref{fig4:topics}{Figure}{A} shows that this achievement does not translate to the OpenAlex datasets where the embedding model AllMPNetBase achieves a higher success rate. 

AllMPNetBase is a monolingual embedding model, so would not be expected to perform well on Wikipedia in other languages, although it does achieve a reasonable success rate in French.
The vast majority of the OpenAlex articles are written in English, but the success rate of AllMPNetBase over E5-Small for Biology, Medicine and Physics is significant, suggesting they have different representations of scientific language.
The results demonstrate, that the relative performance of embedding models on one dataset will not translate to the context of another dataset.

\subsection{LLMs}

\begin{figure}[t!]
    \centering
    \includegraphics[width=\linewidth]{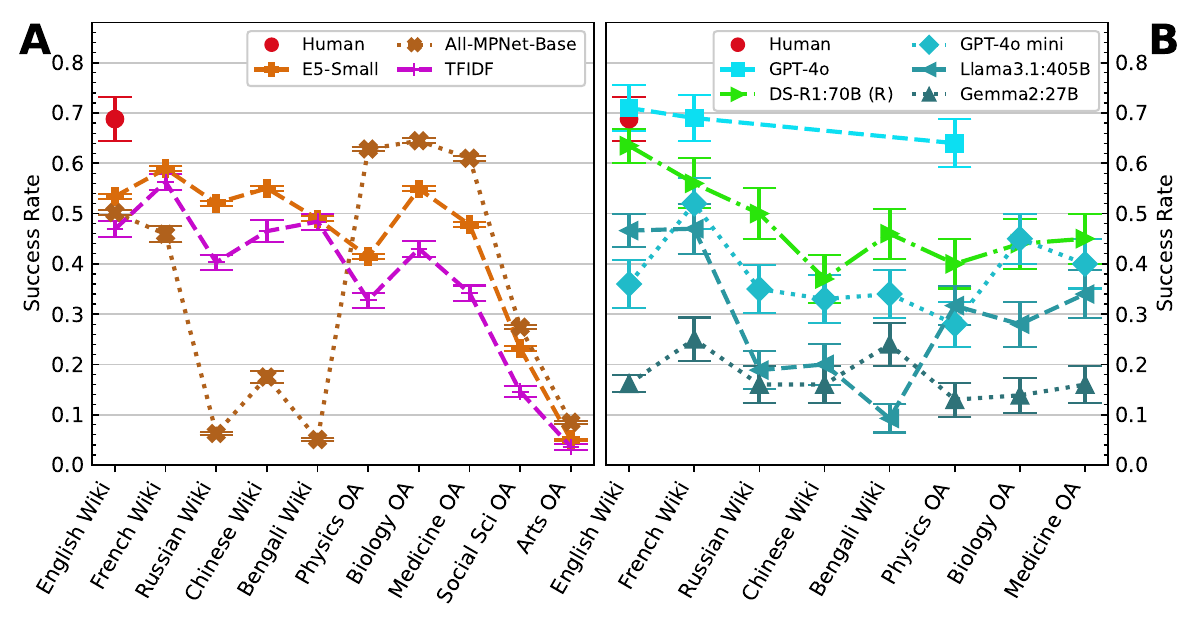}
    \caption{\protect\input{figures/fig4_caption}}
    \label{fig4:topics}
\end{figure}

\linkref{fig4:topics}{Figure}{B} shows the success rate of LLMs on eight of the graph datasets (the LLMs were not applied to the OpenAlex Social Sciences and Arts datasets). For the LLMs, the success rate is ordered roughly by model size but is not consistent between topics.
The GPT-4o models are mostly consistent across the range of languages and topics, with the GPT-4o mini's performance not statistically significantly dropping below the performance on English Wikipedia.

The success rate of Llama:405B deteriorated when operating in many languages and topics outside of English Wikipedia.
When tested on the OpenAlex dataset, the performance of the 405~billion-parameter model dropped by around a third, and the 3~billion-parameter Llama model had a success rate near or at zero.
This suggests that, as with the embedding models, general aptitude might not transfer between subject areas.

\section{Discussion}
\label{section_discussion}
\typeout{~~ Discussion ~~}

This report shows that sentence embedding models and LLMs can contain sufficient semantic information to play the Wikipedia Game successfully.
This is not necessarily a surprising result, as English Wikipedia is ubiquitous across NLP training datasets.
Still, particularly with LLMs, a gulf can exist between being able to recite a fact and being able to utilise it~\cite{nezhurinaAliceWonderlandSimple2024}.

While the best-performing models are able to achieve a comparable success rate to humans on English Wikipedia, it took more steps on average, demonstrating that it is still not as efficient as humans at this task.
This differs from what LLM-WikiRace~\cite{ziomekLLMWikiRaceBenchmarkHow2026} found, but that difference likely stems from evaluating on an `easy' set of games where humans fail only 1.5\% of the time.
Other work, as expected, found that models gain efficiency compared to humans when backtracking is allowed~\cite{westAutomaticHumanNavigation2012}.
The time limit likely plays a part in this, but it will also be humans' reluctance to balance many different options like an embedding model does automatically.
However, the LLMs were more efficient than the embedding agents and this may indicate that they are leveraging existing knowledge of the Wikipedia Game (shown in \supLLMExplain) to influence the selection of links.
\supRoutes shows GPT-4o making some beneficial choices in the early steps of the game to articles that the embedding models would not consider to be any closer to the target than the starting article.

The results unexpectedly showed Llama:8B outperformed Llama:70B, which suggests these smaller models' training process might not have been perfectly optimised.
The two largest models, GPT-4o and Llama:405b, showed a significant difference in success rate contrary to \cite{grattafioriLlama3Herd2024}, which concludes they have comparable quality on a range of tasks.
Another study that compared LLMs concluded that GPT-4o has superior performance when the context is long and that Llama:405b is best for short contexts \cite{maekawaHolisticReasoningLongContext2024}.
The requirement to consider many possible next steps may explain the difference in these models' performance on the Wikipedia Game.
This agrees with models trained for reasoning outperforming similarly sized non-reasoning models~\cite{ziomekLLMWikiRaceBenchmarkHow2026}; the QwQ:32B performed comparably to GPT-4o.
The performance increase was not seen when using conversational models with a reasoning technique.
The difference in efficiency was still stark when compared to humans, and the step count distribution strongly suggests that there are big differences between language models and humans in how they play the game.

This work also investigated whether performance would generalise to different languages and datasets, as, in other tasks, deficiencies are seen when transitioning to a lower-resource environment~\cite{hasanLargeLanguageModels2024}.
The results in \linkref{section_lang_topics}{Section}{} showed that there can be significant differences in performance by language and topic. 
The features of each graph dataset will also impact the success rate, with smaller graphs potentially being easier to navigate.
For example, some models have a higher success rate on French Wikipedia than they do in English.
This might reflect the fact that the French Wikipedia graph is both smaller and better connected rather than the models necessarily being better at French.

Many characteristics of the OpenAlex datasets are different from English Wikipedia, with fewer links per article and longer titles with more specialist language.
Despite this, the success rates for some of the LLMs on the Biology, Medicine and Physics datasets were within the margin of error of the success rate on the English Wikipedia dataset.
This indicates that the LLMs are not simply recalling links that they have already `learned' during training on Wikipedia data but have some ability to generalise the concept of the game to a different dataset.
One might wonder whether the models can still navigate to articles that were published beyond the training date.
This was explored with a very small sample of target Wikipedia articles published in 2024, after the models were trained (\supPostTraining). 
These articles are less well connected, so intrinsically more difficult to reach, but every model's success rate fell proportionally less than TF-IDF's.

There are some natural extensions to this work; for example, one could apply the large GPT-4o model and gather more benchmark data from humans playing the game on all datasets.
This would allow insights between languages not possible from the TF-IDF technique.
For example, the Bengali Wikipedia Game might be extremely easy for a native speaker owing to the small size of the graph.
Moreover, many native English speakers might struggle to navigate OpenAlex datasets owing to the specialised nature of the language.

This work focused on LLMs and sentence embedding models, but other techniques could be interesting to consider.
Cross-encoder models typically have a computational cost somewhere between the embedding and the LLM approaches used in this work.
Other potential approaches include hybrid methods, such as presenting only the top choices from the embedding model in an LLM prompt or using RL to fine-tune an existing model.
Now that this method of challenging NLP models has been established, we look forward to seeing how more complex techniques can be applied.

\section{Conclusions}
\typeout{~~ Conclusions ~~}

This project has investigated whether sentence embedding models and LLMs contain sufficient semantic information about related concepts to navigate an interconnected graph of articles.
26 different language models are tested on 10 graph datasets generated from Wikipedia and OpenAlex.
This experimental set-up resembled a real-world task, motivating applicability and allowing comparison with a human benchmark on the English Wikipedia dataset.

The performance of the models varied greatly on type, size, and testing domain.
Notably, the best LLMs used in the experimental work achieved a comparable success rate to humans, though they took more steps on average than a human.
Performance deteriorated with decreasing size of the LLM, and hallucinations often negatively impacted their performance. 
By contrast, the sentence embedding models do not suffer from hallucinations and some were able to outperform many LLMs, achieving higher success rates.
Reasoning models achieved much better performance for their size, but their efficiency more closely matched the non-reasoning and embedding models.
It is important to note that running a reasoning LLM is vastly more computationally expensive than an embedding model, and in many practical circumstances the simpler approach would be optimal.

The results showed that the models' relative performance was not consistent between topics as their ranking changed between languages and academic fields.
While none of the models significantly exceeded human performance on the English Wikipedia dataset, this result would not necessarily translate to the more technical datasets, such as Biology, Medicine, or Physics, where specialist knowledge would be required.
Furthermore, given the relative time and monetary costs of human processing compared to language models, there will be practical tasks where these models are faster and cheaper than human labour, though this work shows to achieve this care must be taken when applying these techniques across domains.

\section*{Acknowledgements}
The computations described in this research were performed using the Baskerville Tier 2 HPC service. Baskerville was funded by the EPSRC and UKRI through the World Class Labs scheme (EP/T022221/1) and the Digital Research Infrastructure programme (EP/W032244/1) and is operated by Advanced Research Computing at the University of Birmingham.

\bibliographystyle{splncs04}
\bibliography{92WikipediaGame}

\clearpage
\appendix

\section{Wikipedia Data}\label{appendix_wikipedia}
\typeout{~~ Wikipedia Data ~~}

\subsection{Access to Data}

Wikipedia has an API, but due to limited server capacity, they request that for use cases like ours, we download the data dumps of the website.\footnote{WikiMedia API Main Page - \url{https://www.mediawiki.org/wiki/API:Main_page}}
Multiple versions of the data dumps contain different information in different formats.\footnote{Wikipedia data dumps  - \url{https://dumps.wikimedia.org/enwiki/latest/}}
These data sources are split into two main groups.
The database contains the meta-information about the pages, and unique IDs for them, and where each page links.
The data dumps are a large file or a collection of files that contain the Wiki markup which makes up the article and a small amount of associated metadata to identify the article.

\subsection{Page Links and the Database}

The database containing the metadata about the articles on Wikipedia has multiple tables, which contain metadata about all the pages and the links which the pages contain.
Using this database comes with one important consideration, which is that the list of links also includes all the navigation links at the bottom of the page.
These links can be included in the Wikipedia game depending on the rules, but it will be very difficult to make a language model that can consider them as they are not included in the article markup.
A lot of Wikipedia articles will have the majority of their links included in collapsed lists of links at the bottom of the page, which are unlikely to be parseable by a language model without extra processing.

\subsection{The Article Dumps}

All of the text of Wikipedia is available from the Wikipedia dumps, and also the Talk pages and edit history if required.
For our purposes, the current version of the text of the articles would be the most useful, and this information is included without much else in a dump called `latest-pages-articles`.
This dump is formatted in XML with some extra information as well as the text, including the title of the page and whether or not it is a redirect page.
The XML format is reasonably minimal with tags such as \texttt{<page>} and \texttt{<title>} segmenting the data.
There are also open-source wiki-XML parsers available, including \texttt{python-mwxml}\footnote{Python MediaWiki - \url{https://github.com/mediawiki-utilities/python-mwxml}} and \texttt{mwparserfromhell}\footnote{The Wikimedia Parser from Hell - \url{https://github.com/earwig/mwparserfromhell}}.

As Wikipedia is very large if all the articles are included in one file that file ends up totalling 90GB which introduces difficulties when it comes to processing it.
To help with this the articles are available split up into twenty-seven smaller XML files, and an index file which lists the articles in the order that they appear in the XML documents.
The XML files are also available in a compressed format, which can be useful but we had enough storage space to not require this.

\subsection{Producing the Graph}

The graph must align with the summary data; therefore, only the article dumps were utilized to construct the graph.
This approach minimized any discrepancies between the database and the article information based on the time the snapshot was produced. 
The title, summary, and links within the article were all extracted from the Wikipedia markup. Additionally, the pages include a flag that indicates whether they are `redirect' pages.
This feature allows Wikipedia to link a single page to multiple aliases. 
For instance, the page `Accessible computing' redirects to `Computer accessibility', meaning that either of these titles can be used as links and will direct users to the same content.

Once these data are collected, all the article pages are indexed, and the links can be stored by following through the redirects to create a list of article index pairs.
These indexes, corresponding to each article, serve as the nodes of the graph, while the links provide the edges.
A directed graph is formed because not every linked page will necessarily contain a link to return to the original page. 

The titles and summaries of the articles are stored alongside their respective indexes.
Throughout Wikipedia, the titles are case-sensitive apart from the first character, and for this work, we capitalised the first letter of every title.
The summaries had the WikiMedia markup stripped away, and the first 1,000 characters of the main text were used for the summaries.
For the navigation task, the title of the article was given to the language model as opposed to any redirect that might have been linked to the current page.

\clearpage

\section{OpenAlex Data}\label{appendix_openalex}
\typeout{~~ OpenAlex Data ~~}

\subsection{About OpenAlex}

OpenAlex is a very large collection of metadata about academic literature named after the Library of Alexandria.
It is an open-source and openly available collection with over 200 million pieces of academic work.
Information about the work is organised in a relational database structure which also contains information about authors and institutions among others.
This dataset is constructed by assembling information from multiple sources including the Microsoft Academic Graph\footnote{Microsoft Academic - \url{https://www.microsoft.com/en-us/research/project/academic/}} and Crossref.\footnote{Crossref - \url{https://www.crossref.org/}}

The most important part of OpenAlex for our application is the `Works' database which contains the metadata about the pieces of academic literature.
The metadata we have made use of is the title, abstract, referenced works, language, primary topic, and the type of the article.
The DOI for each article is also stored for debugging processes.
Only articles which had valid data for all of these fields were used in the dataset that we constructed. 

The title, abstract, and type of article are extracted from the metadata from the datasets that OpenAlex is built from, which means they are only available if they are specified.
The language and the primary topic are inferred by OpenAlex from the other data available, which means that it is not perfectly reliable.
For example, a medical study with lots of Latin names interwoven in the English text can be misclassified as French or Italian.\footnote{Article in English misclassified - \url{https://api.openalex.org/works/W2412377776}}

\subsection{Access to the Data}

OpenAlex has an API\footnote{OpenAlex API - \url{https://docs.openalex.org/how-to-use-the-api/api-overview}} for querying information but with the amount of information required to confirm if two articles are connected, we required an offline version of the graph.
The complete OpenAlex dataset is available from an Amazon Web Services storage location which can then be processed locally.\footnote{OpenAlex Snapshot - \url{https://docs.openalex.org/download-all-data/openalex-snapshot}}
This dataset totals 1.6~TB when in its uncompressed format so thought needs to be taken when extracting the data.
The approach we took is to uncompress it in stages and then discard the uncompressed data, preserving only the components that we need.

The dataset is being continuously updated and unlike Wikipedia historical snapshots are not available.
This means that the experiments included in this report would be slightly different if run again on the most recent version of OpenAlex.

\subsection{Building the Graph}

The process to build the graph resembled that for Wikipedia, but substituting pages for scientific articles and links for citations.
To explore the effect of topic and to bring the dataset to a size comparable to that of Wikipedia, we filtered the primary topic.
Five different fields of research were considered Physics and Astronomy; Biochemistry, Genetics, and Molecular Biology; Medicine; Social Sciences; and Arts and Humanities.

Of the entries in the OpenAlex works database only the type `article' was used, which excluded book chapters among others.
Once this list of articles was generated for the topic of interest then they were indexed in order to construct the graph.
The OpenAlex articles already have an index (work id) provided but as graph path solving requires matrix operations making the total number smaller provides a computational speed up.
The work id of the article was also stored with the metadata which was useful for exploratory purposes but was not used by the agents.

The edges of the graph are given by the citation network, all articles that were cited by an article or that cited the article were navigable to.
This choice to make the graph symmetric was necessary, as if only the cited articles were used, then it would only be possible to go backwards by publication date.
This alternative way of setting up the game would be extraordinarily difficult, as a game would be lost as soon as an article was chosen with a date older than the target.
Given that it is possible to navigate both ways through every citation the graph produced for the OpenAlex works is symmetric, unlike the Wikipedia graph.
Despite this it is still possible for the agents to reach a dead end because they are forbidden from navigating to already visited articles to stop them becoming trapped in loops.

The abstracts in OpenAlex were used in a comparable way to the summaries in the Wikipedia dataset.
Before the abstracts could be used they needed to be converted from the inverted index format that they were stored in.
Unfortunately some of the titles and abstracts had formatting information stored in them and this was not trivial to remove as it was not in a consistent format like the WikiMedia markup in the Wikipedia summaries.
This only applied to a small number of articles though, and was most common in the Physics dataset where the abstract would contain equations or mathematical terms.
\clearpage

\section{Graph Statistics}\label{appendix_datasets}
\typeout{~~ Graph Statistics ~~}

\linkref{tab:dataset_summary}{Table}{} lists the datasets used in this work. Some of the OpenAlex disciplines have been given short names: `Biochemistry, Genetics and Molecular Biology', `Physics and Astronomy' and 'Arts and Humanities' have been shortened to `Biology', `Physics' and `Arts', respectively.

The table shows the percentage of nodes (articles) that are disconnected, unreachable and dead-ends. Disconnected nodes have no links in or out of them. Unreachable nodes have links going out to other nodes, but no links into the node - these could be used as the starting node for a game but will never feature as a target node or in the route taken during a game. Dead-end nodes have links into them, but no links out. If these are navigated to during the course of a game then the game will end as there will be no node that it will be possible to move to next. 

The OpenAlex datasets have a high percentage of disconnected nodes compared to the Wikipedia datasets, which was also reflected in \linkref{fig1:graph}{Figure}{B} in the percentage of nodes that can connect. In particular, over half of the nodes for the OpenAlex Social Sciences and Arts datasets are disconnected. 

The table also lists the mean number of edges per node and per connected node (i.e., all nodes that are not disconnected). As well as having a high percentage of disconnected nodes, the Social Sciences and Arts datasets have a relatively low mean number of edges per node showing that the articles in these datasets are not as well connected as other datasets. 

The mean number of characters used in the titles of each dataset are also displayed in the table. These show that the OpenAlex datasets have longer article titles than the Wikipedia datasets.

\begin{landscape}
\begin{table}[ht]
\begin{adjustbox}{width=1.4\textwidth}
\centering
\renewcommand{\arraystretch}{1.5}
\begin{tabular}{ccccccccccc}
 \multirow{2}{*}{\textbf{Source}} & \multirow{2}{*}{\textbf{Language}} & \multirow{2}{*}{\textbf{Discipline}} & \multirowcell{2}{\textbf{Number}\\\textbf{of Nodes}} & \multirowcell{2}{\textbf{Number}\\\textbf{of Edges}} & \multicolumn{3}{c}{\textbf{Percentage of Nodes}} & \multirowcell{2}{\textbf{Mean Edges}\\\textbf{per Node}} & \multirowcell{2}{\textbf{Mean Edges per}\\\textbf{Connected Node}} & \multirowcell{2}{\textbf{Mean Length}\\\textbf{of Titles}}\\ 
  & & & & & \textbf{Disconnected} & \textbf{Unreachable} & \textbf{Dead-ends} &\\ \hline
 Wikipedia & Bengali & Multiple & 158,000 & 3,152,000 & 1.43\% & 23.77\% & 1.01\% & 19.97 & 20.26 & 17.55\\
 Wikipedia & Chinese & Multiple & 1,437,000 & 36,785,000 & 0.73\% & 22.84\% & 2.40\% & 25.60 & 25.79 & 6.45\\
 Wikipedia & English & Multiple & 6,871,000 & 295,749,000 & 0.11\% & 8.23\% & 0.32\% & 43.04 & 43.09 & 19.89\\
 Wikipedia & French & Multiple & 2,630,000 & 117,860,000 & 0.09\% & 7.70\% & 0.06\% & 44.81 & 44.85 & 19.14\\
 Wikipedia & Russian & Multiple & 1,997,000 & 67,546,000 & 0.40\% & 11.36\% & 0.17\% & 33.82 & 33.96 & 20.32\\ \hdashline
 OpenAlex & Multiple & Arts & 3,987,000 & 7,429,000 & 70.05\% & 0.00\% & 0.00\% & 1.86 & 6.22 & 78.27\\
OpenAlex & Multiple & Biology & 6,569,000 & 138,492,000 & 25.67\% & 0.00\% & 0.00\% & 21.08 & 28.37 & 100.01\\
 OpenAlex & Multiple & Medicine & 20,157,000 & 334,463,000 & 31.59\% & 0.00\% & 0.00\% & 16.59 & 24.26 & 97.05\\
 OpenAlex & Multiple & Physics & 4,199,000 & 115,747,000 & 19.17\% & 0.00\% & 0.00\% & 27.56 & 34.10 & 87.21\\
 OpenAlex & Multiple & Social Sciences & 12,570,000 & 72,936,000 & 57.64\% & 0.00\% & 0.00\% & 5.80 & 13.70 & 84.74\\
\hline
\end{tabular}
\end{adjustbox}
\caption{The datasets used in this project. The nodes in the dataset are pages in Wikipedia and articles in OpenAlex. The edges are links in Wikipedia and references and citations in OpenAlex. The number of nodes and edges have been rounded to the nearest 1,000.}
\label{tab:dataset_summary}
\end{table}
\end{landscape}
\clearpage

\section{TF-IDF}\label{appendix_tfidf}
\typeout{~~ TF-IDF ~~}

The term frequency-inverse document frequency (TF-IDF) approach requires a corpus of documents.
The corpus is used to create a vocabulary of words and the IDF value for each word in the vocabulary.
The IDF for a word (or term) $t$ and corpus $D$ is calculated as:
\begin{equation}
    \text{IDF}(t, D) = \log\frac{N}{n_t}
\end{equation}
where $N$ is the total number of documents in the corpus and $n_t$ is the number of documents where the term $t$ appears.

For any given document $d$, term $t$, the TF is calculated as:
\begin{equation}
    \text{TF}(t, d) = \frac{f_{t,d}}{\sum_{t^{'}\in d}{f_{t^{'},d}}}
\end{equation}
where $f_{t,d}$ is the count of term $t$ in document $d$, and the denominator is the total number of terms in $d$.

The TF-IDF value for a term $t$, document $d$ and corpus $D$ is then given as:
\begin{equation}
    \text{TF-IDF}(t,d,D) =  \text{TF}(t,d)\cdot\text{IDF}(t,d,D)
\end{equation}

The TF-IDF agent plays the Wikipedia Game in a manner similar to that of the sentence embedding agent.
It compares the TF-IDF vectors for the target article and the possible reachable articles, using cosine distance, choosing the article with minimum distance to the target.
\clearpage

\section{Embedding Models}\label{appendix_embedding_models}
\typeout{~~ Embedding Models ~~}

\linkref{tab:embedding_models}{Table}{} lists the sentence embedding model  used in the experiments. A `short name' for each model is also provided; these will be used in subsequent tables in the appendices. 
\begin{table}[ht]
\begin{adjustbox}{width=1.0\textwidth}
\centering
\renewcommand{\arraystretch}{1.5}
\begin{tabular}{cccc}
\textbf{Sentence Embedding Model} & \textbf{Short Name} & \textbf{Parameters (Millions)} & \textbf{Multilingual}  \\
\hline
\href{https://huggingface.co/intfloat/multilingual-e5-large}{multilingual-e5-large} & E5-Large & 560 & Y \\
\href{https://huggingface.co/sentence-transformers/all-roberta-large-v1}{all-Roberta-large-V1} & Roberta & 355 & N \\
\href{https://huggingface.co/sentence-transformers/stsb-xlm-r-multilingual}{stsb-xlm-r-multilingual} & STSB-XLM-R & 278 & Y \\
\href{https://huggingface.co/sentence-transformers/paraphrase-multilingual-mpnet-base-v2}{paraphrase-multilingual-mpnet-base-v2} & Para-MPNet-Base & 278 & Y \\
\href{https://huggingface.co/sentence-transformers/distiluse-base-multilingual-cased-v2}{distiluse-base-multilingual-cased-v2} & Distiluse-Base-v2 & 135 & Y \\
\href{https://huggingface.co/sentence-transformers/distiluse-base-multilingual-cased-v1}{distiluse-base-multilingual-cased-v1} & Distiluse-Base-v1 & 135 & N \\
\href{https://huggingface.co/sentence-transformers/average_word_embeddings_glove.6B.300d}{average\_word\_embeddings\_glove.6B.300d} & Glove & 120 & N \\
\href{https://huggingface.co/intfloat/multilingual-e5-small}{multilingual-e5-small} & E5-Small & 118 & Y \\
\href{https://huggingface.co/sentence-transformers/paraphrase-multilingual-MiniLM-L12-v2}{paraphrase-multilingual-MiniLM-L12-v2}& MiniLM-L12 & 118 & Y \\
\href{https://huggingface.co/sentence-transformers/all-mpnet-base-v2}{all-mpnet-base-v2} & AllMPNetBase & 109 & N \\
\href{https://huggingface.co/sentence-transformers/all-distilroberta-v1}{all-distilroberta-v1} & DistilRoberta & 82.1 & N \\
\href{https://huggingface.co/sentence-transformers/all-MiniLM-L6-v2}{all-MiniLM-L6-v2} & MiniLM-L6 & 22.7 & N \\
\href{https://huggingface.co/sentence-transformers/paraphrase-albert-small-v2}{paraphrase-albert-small-v2} & Albert-Small & 11.7 & N \\
\end{tabular}
\end{adjustbox}
\caption{The sentence embedding model used in this project, the number of parameters in each model and whether or not the model is trained on multilingual data. Each model is linked to its HuggingFace record.}
\label{tab:embedding_models}
\end{table}

\clearpage

\section{LLM Models}\label{appendix_llm_models}
\typeout{~~ LLM Models ~~}

\linkref{tab:llms}{Table}{} lists the prompt-based LLM models used in the experiments.
Mistral NeMo was only run on one dataset and is not discussed in the main text.
The size of the GPT-4o models has not been released by OpenAI and reliable estimations have not been made. The best estimations that were used to create \linkref{fig2:performance}{Figure}{} have been documented on Wikipedia\footnote{Size of GPT-4 - \url{https://en.wikipedia.org/wiki/GPT-4}}, the Decoder\footnote{The decoder - \url{https://the-decoder.com/gpt-4-architecture-datasets-costs-and-more-leaked/}}, TechCrunch\footnote{TechCrunch - \url{https://techcrunch.com/2024/07/18/openai-unveils-gpt-4o-mini-a-small-ai-model-powering-chatgpt/}} and Reddit\footnote{Reddit LocalLlama - \url{https://www.reddit.com/r/LocalLLaMA/comments/1f1vpyt/why_gpt_4o_mini_is_probably_around_8b_active/}}
\begin{table}[ht]
\centering
\renewcommand{\arraystretch}{1.5}
\begin{tabular}{ccc}
\textbf{Model} & \textbf{Parameters (Billions)} & \textbf{Size (GB)} \\
\hline
\multirow{2}{*}{Llama3.2} & 1 & 2 \\
 & 3 & 3 \\ \hdashline
\multirow{3}{*}{Llama3.1} & 8 & 5 \\
 & 70 & 40 \\
 & 405 & 230 \\ \hdashline
Gemma2 & 27 & 16 \\ \hdashline
Mistral NeMo & 12 & 7 \\ \hdashline
\multirow{3}{*}{DeepSeek-R1} & 8 & 5 \\
   & 32 & 20 \\
   & 70 & 43 \\ \hdashline
QwQ & 32 & 20 \\ \hdashline
GPT-o3 mini & Unknown & Unknown \\ \hdashline
GPT-4o mini & Unknown & Unknown \\
GPT-4o & Unknown & Unknown \\
\end{tabular}
\caption{
    The generative LLM used in this project, the number of parameters in each model (if known) and the size of the model in GB (if known).
}
\label{tab:llms}
\end{table}
\clearpage

\section{LLMs and the Wikipedia Game}\label{appendix_llm_wiki}
\typeout{~~ LLMs and the Wikipedia Game ~~}

The listings in this section show the response from some of the LLMs used in this project when given the following prompt: `Do you know how to play the Wikipedia game?'. \linkref{lst_gpt_4o_wiki_prompt}{Listing}{} shows the response from GPT-4o, which demonstrates that it is able to explain the rules of the game. Similarly, \linkref{lst_gemma_wiki_prompt}{Listing}{} and \linkref{lst_llama_8b_wiki_prompt}{Listing}{} provide the responses from Gemma:27b and Llama:8b and both are able to explain the game. However, the smaller models were not able to explain the game and \linkref{lst_llama_3b_wiki_prompt}{Listing}{} shows the response from Llama:3b. 

\begin{lstlisting}[breaklines,basicstyle=\ttfamily,caption=The response from GPT-4o when it was asked if it knew how to play the Wikipedia Game,label=lst_gpt_4o_wiki_prompt]
    Yes, I'm familiar with the Wikipedia Game! It's a fun and educational game that involves navigating from one Wikipedia page to another using only the links within the articles. Here's how you can play it:
Basic Rules:

    Choose the Starting and Ending Articles:
        Starting Article: Begin on a specific Wikipedia article.
        Ending Article: Decide on a target article that you need to reach.
    Navigate Only Through Links:
        You can only click on the hyperlinks within the articles to move from one page to another.
        You cannot use the search bar or any external links.
    Objective:
        The goal is to reach the ending article from the starting article in the fewest number of clicks or in the shortest amount of time.

Variations:

    Speed Race: Compete with others to see who can reach the target article the fastest.
    Fewest Clicks: Compete to see who can reach the target article in the fewest number of clicks.

Example:

    Starting Article: "Butterfly"
    Ending Article: "Albert Einstein"

    Your task would be to navigate from the "Butterfly" article to the "Albert Einstein" article by clicking on links within the articles. You might go from "Butterfly" to "Insect," then to "Biology," and so on, until you reach "Albert Einstein."

Tips for Playing:

    Look for broad topics that can lead you closer to your target.
    Use categories at the bottom of the articles to find broader or related topics.
    Keep track of the articles you've visited to avoid going in circles.

    It's a great way to learn about different topics and see how interconnected information can be. Enjoy playing the Wikipedia Game!

\end{lstlisting}

\begin{lstlisting}[breaklines,basicstyle=\ttfamily,caption=The response from Gemma:27b when it was asked if it knew how to play the Wikipedia Game,label=lst_gemma_wiki_prompt]
    Yes, I do!
    
    The Wikipedia Game is a fun way to test your knowledge and explore the vast interconnectedness of Wikipedia. Here's how it works:
    
    **The Rules:**
    
    1. **Starting Point:** You begin with a randomly chosen Wikipedia article. 
    2. **Target:** You aim to reach a predetermined target article, usually something very specific like "Philosophy" or "Kevin Bacon".
    3. **Moves:** On each turn, you can only click on links within the current Wikipedia article to navigate to another page.
    
    **Winning:** The first person to reach the target article wins!
    
    **Variations:**
    
    * Some people set a time limit for the game.
    * Others introduce restrictions, like forbidding the use of the search bar.
    
    **Do you want to play? I can help by suggesting starting articles or target articles if you'd like.
\end{lstlisting}

\begin{lstlisting}[breaklines,basicstyle=\ttfamily,caption=The response from Llama:8b when it was asked if it knew how to play the Wikipedia Game,label=lst_llama_8b_wiki_prompt]
    The Wikipedia Game is a popular online game that involves navigating from one Wikipedia article to another using only hyperlinks within the articles themselves.
    
    To play, I'll need some initial guidance. Here's how it usually works:
    
    1. We start with a random Wikipedia article.
    2. The goal is to reach a predetermined target article by clicking on links in each article and following them to related topics.
    3. Each link must be clicked exactly once; we can't go back or revisit any articles already visited.
    
    Sounds like fun? Do you want to play, and if so, would you like me to start with a random article, or do you have a specific one in mind as the starting point?
    
    Also, please let me know what kind of difficulty level you'd prefer: easy (e.g., navigating between articles on a similar topic), medium (jumping between somewhat related topics), or challenging (moving from very diverse subjects).
\end{lstlisting}

\begin{lstlisting}[breaklines,basicstyle=\ttfamily,caption=The response from Llama:3b when it was asked if it knew how to play the Wikipedia Game,label=lst_llama_3b_wiki_prompt]
    I do know how to play the Wikipedia Game, also known as Wiki-Trivia or Random Wikipedia. Here's a brief overview of how we can play:
    
    1. I'll think of a random topic and provide it to you.
    2. You try to come up with a question related to that topic.
    3. You ask your question on me, and I'll respond with the answer.
    4. If you're correct, I'll give you another question based on the same topic. If not, we move on to a new topic.
    
    Ready to get started? Would you like a randomly generated topic or would you like to choose one yourself?
    
    (Also, please keep in mind that the Wikipedia Game is all about learning and having fun, so don't worry if you don't know much about a particular topic. I'll be happy to help and provide more information as needed!)
\end{lstlisting}
\clearpage

\section{LLM Prompt}\label{appendix_llm_prompt}
\typeout{~~ LLM Prompt ~~}

The code shown in \hyperref[lst_main_prompt]{Listing \ref{lst_main_prompt}} was used for the prompt for the start of each round of the Wikipedia game for all LLMs. Note that the prompt does not contain any information about the current article, other than the titles of the articles that are linked on the current article. As described in the main text, the code has a configurable setting allowing a number of retries if the LLM fails to return a valid answer. In the case of a failure, the response is added to the prompt for the round and an additional message is appended, depending on the reason for the failure. Four failure reasons are catered for, which each have a different prompt which are shown in the following listings:
\begin{itemize}
    \item A blank or null response, \hyperref[lst_none_output]{Listing \ref{lst_none_output}}
    \item The response is not in a valid JSON format, \hyperref[lst_not_json]{Listing \ref{lst_not_json}}
    \item The response is in JSON format, but does not contain the key `title', \hyperref[lst_json_no_title]{Listing \ref{lst_json_no_title}}
    \item The response is in JSON format, contains the key `title', but the given title is a hallucination and does not match one on the current article, \hyperref[lst_hallucination]{Listing \ref{lst_hallucination}}
\end{itemize}

\begin{lstlisting}[breaklines,basicstyle=\ttfamily,caption=The prompt given at the beginning of each round (step) of the game where <target\_title> <target\_summary> and <list\_of\_titles> will be replaced by the values for the given round of the game., label=lst_main_prompt]
    {"role": "system",
     "content": "You are a helpful assistant that helps me to play the Wikipedia Game! "},
     
    {"role": "user",
     "content": "If I give you the title and a summary of the target page that I am trying to navigate to, can you help me navigate to the target page? "},
     
    {"role": "assistant",
     "content": "Of course! What is your target page?"},
     
    {"role": "user",
     "content": "{`title': `<target_title>', `summary': `<target_summary>'}},

    {"role": "assistant",
     "content": "Sure! I can help you navigate to that page, what titles are there on your current page?"}

    {"role": "user",
     "content": "Here is a list of titles, which one should I pick? Please provide your answer in JSON format with the selected title given with key `title' and a justification given with key `reason'. The selected title must be from the following list: <list_of_titles>"}
     
\end{lstlisting}

\begin{lstlisting}[breaklines,basicstyle=\ttfamily,caption=The additional message added to the prompt if a blank output is generated by the LLM (this might happen if a content filter prevents a response),label=lst_none_output]

    {"role": "system",
     "content": "{`title': `Unknown title', `reason': `I did not know which title to select.'}"

    {"role": "user",
     "content": "But `Unknown title' is not in the list that was provided! You must select a title from the list of titles. Please try again and provide your answer in JSON format with the selected title given with key `title' and a justification given with key `reason'. The JSON should only include the two fields `title' and `reason', and the title should only include the title text. You are tying to select the best page title to reach the target page `<target_title>'."}
    
\end{lstlisting}

\begin{lstlisting}[breaklines,basicstyle=\ttfamily,caption=The additional message added to the prompt if the output generated by the LLM is not in JSON format,label=lst_not_json]
    {"role": "user",
     "content": "You did not provide your response in JSON format - please try again! The JSON should only include the two fields `title' and `reason', and the title should only include the title text. You are tying to select the best page title to reach the target page `<target_title>'."
\end{lstlisting}

\begin{lstlisting}[breaklines,basicstyle=\ttfamily,caption=The additional message added to the prompt if the output generated by the LLM is in JSON format but does not contain the key `title',label=lst_json_no_title]
    {"role": "user",
     "content": "You must provide a title to navigate to. Please try again and provide your answer in JSON format with the selected title given with key `title' and a justification given with key `reason'. The JSON should only include the two fields `title' and `reason', and the title should only include the title text. You are tying to select the best page title to reach the target page `<target_title>'"
\end{lstlisting}

\begin{lstlisting}[breaklines,basicstyle=\ttfamily,caption=The additional message added to the prompt if the output generated by the LLM is in JSON format but contains a hallucination in the `title' field,label=lst_hallucination]
    {"role": "user",
     "content": "But `<title>' is not in the list that was provided! You must select a title from the list of titles. Please try again and provide your answer in JSON format with the selected title given with key `title' and a justification given with key `reason'. To jog your memory, 10 example titles from the list are: <list_of_10_random_titles>. The JSON should only include the two fields `title' and `reason', and the title should only include the title text. You are tying to select the best page title to reach the target page `<target_title>'."}
\end{lstlisting}
\clearpage

\section{Model Performance}\label{appendix_model_performance}
\typeout{~~ Model Performance ~~}

\linkref{tab:model_performance_success_rate}{Table}{} lists the success rate for the TF-IDF, embedding and LLM prompt models used in the experiments. The highest success rate for each of the datasets (within the bounds of error) are highlighted in bold. \linkref{tab:model_performance_mean_steps}{Table}{} lists the mean number of steps above optimal for each of the models. Where the success rate is less than or equal to 0.05, the mean number of steps is not shown. The lowest number of steps (within the bounds of error) for each dataset is highlighted in bold. 
For both tables, the errors are shown as one standard deviation of uncertainty on the predicted fraction. The error bounds will be larger when fewer games have been played with a model.

\begin{landscape}

\begin{table}[ht]
\centering
\renewcommand{\arraystretch}{1.2}
\begin{adjustbox}{width=1.5\textwidth}
\begin{tabular}{l|ccccc|ccccc}
\textbf{} & \multicolumn{5}{c|}{\textbf{Wikipedia}}& \multicolumn{5}{c}{\textbf{OpenAlex}} \\
\textbf{Model} & \textbf{Bengali} & \textbf{Chinese} & \textbf{English} & \textbf{French} & \textbf{Russian} & \textbf{Arts} & \textbf{Biology} & \textbf{Medicine} & \textbf{Physics} & \textbf{Social Sciences} \\ \hline
TF-IDF & 0.48 ± 0.02 & 0.47 ± 0.02 & 0.47 ± 0.02 & 0.56 ± 0.02 & 0.40 ± 0.02 & 0.04 ± 0.01 & 0.43 ± 0.02 & 0.34 ± 0.02 & 0.33 ± 0.01 & 0.15 ± 0.01 \\ \hdashline
AllMPNetBase & 0.05 ± 0.00 & 0.18 ± 0.01 & 0.50 ± 0.00 & 0.46 ± 0.02 & 0.06 ± 0.00 & \textbf{0.09 ± 0.00} & \textbf{0.64 ± 0.00} & \textbf{0.61 ± 0.00} & \textbf{0.63 ± 0.00} & 0.27 ± 0.00 \\
Distiluse-Base-v1 & 0.05 ± 0.00 & 0.34 ± 0.00 & 0.31 ± 0.00 & 0.34 ± 0.00 & 0.24 ± 0.00 & 0.04 ± 0.00 & 0.29 ± 0.00 & 0.25 ± 0.00 & 0.26 ± 0.00 & 0.17 ± 0.00 \\
Distiluse-Base-v2 & 0.03 ± 0.00 & 0.33 ± 0.00 & 0.33 ± 0.00 & 0.41 ± 0.00 & 0.30 ± 0.00 & 0.04 ± 0.00 & 0.30 ± 0.00 & 0.24 ± 0.00 & 0.26 ± 0.00 & 0.17 ± 0.00 \\
E5-Large & \textbf{0.50 ± 0.00 }& 0.53 ± 0.01 & 0.53 ± 0.00 & 0.61 ± 0.00 & \textbf{0.53 ± 0.00} & 0.06 ± 0.00 & 0.61 ± 0.00 & 0.54 ± 0.00 & 0.52 ± 0.00 & 0.25 ± 0.00 \\
E5-Small & 0.49 ± 0.00 & \textbf{0.55 ± 0.00} & 0.53 ± 0.00 & 0.59 ± 0.00 & 0.52 ± 0.00 & 0.05 ± 0.00 & 0.55 ± 0.00 & 0.48 ± 0.00 & 0.41 ± 0.00 & 0.23 ± 0.00 \\
MiniLM-L6 & 0.04 ± 0.00 & 0.11 ± 0.00 & 0.47 ± 0.00 & 0.40 ± 0.00 & 0.05 ± 0.00 & 0.07 ± 0.00 & 0.60 ± 0.00 & 0.56 ± 0.00 & 0.57 ± 0.00 & 0.26 ± 0.00 \\
MiniLM-L12 & 0.08 ± 0.00 & 0.27 ± 0.00 & 0.28 ± 0.00 & 0.34 ± 0.00 & 0.25 ± 0.00 & 0.03 ± 0.00 & 0.37 ± 0.00 & 0.33 ± 0.00 & 0.29 ± 0.00 & 0.15 ± 0.00 \\
Para-MPNet-Base & 0.26 ± 0.00 & 0.26 ± 0.00 & 0.30 ± 0.00 & 0.37 ± 0.00 & 0.26 ± 0.00 & 0.03 ± 0.00 & 0.35 ± 0.00 & 0.31 ± 0.00 & 0.29 ± 0.00 & 0.17 ± 0.00 \\
STSB-XLM-R & 0.22 ± 0.00 & 0.20 ± 0.00 & 0.29 ± 0.00 & 0.35 ± 0.00 & 0.24 ± 0.00 & 0.02 ± 0.00 & 0.15 ± 0.00 & 0.10 ± 0.00 & 0.13 ± 0.00 & 0.12 ± 0.00\\ 
Albert-Small & - & - & 0.28 ± 0.00 & - & - & 0.04 ± 0.00 & 0.44 ± 0.00 & 0.37 ± 0.00 & 0.35 ± 0.00 & 0.18 ± 0.00 \\
DistilRoberta & - & - & 0.44 ± 0.00 & - & - & 0.07 ± 0.00 & 0.60 ± 0.00 & 0.57 ± 0.01 & 0.58 ± 0.00 & 0.26 ± 0.00 \\
Glove & - & - & 0.27 ± 0.00 & - & - & 0.03 ± 0.00 & 0.25 ± 0.00 & 0.17 ± 0.00 & 0.20 ± 0.00 & 0.11 ± 0.00 \\
Roberta & - & - & 0.47 ± 0.00 & - & - & 0.08 ± 0.00 & \textbf{0.64 ± 0.00} & 0.58 ± 0.00 & 0.61 ± 0.00 & \textbf{0.28 ± 0.00} \\\hdashline
GPT-4o mini & 0.34 ± 0.05 & 0.33 ± 0.05 & 0.36 ± 0.05 & 0.52 ± 0.05 & 0.35 ± 0.05 & - & 0.45 ± 0.05 & 0.40 ± 0.05 & 0.28 ± 0.04 & - \\
GPT-4o & - & - & \textbf{0.71 ± 0.05} & \textbf{0.69 ± 0.05} & - & - & - & - & \textbf{0.64 ± 0.05} & - \\ \hdashline
DS-R1:70B (R) & 0.46 ± 0.05 & 0.37 ± 0.05 & 0.64 ± 0.03 & 0.56 ± 0.05 & 0.50 ± 0.05 & - & 0.44 ± 0.05 & 0.45 ± 0.05 & 0.40 ± 0.05 & - \\ \hdashline
Gemma:27b & 0.24 ± 0.04 & 0.16 ± 0.04 & 0.16 ± 0.02 & 0.25 ± 0.04 & 0.16 ± 0.04 & - & 0.14 ± 0.03 & 0.16 ± 0.04 & 0.13 ± 0.03 & - \\
Llama:1b & - & - & 0.01 ± 0.00 & - & - & - & - & - & - & - \\
Llama:3b & 0.09 ± 0.03 & 0.02 ± 0.01 & 0.08 ± 0.01 & 0.07 ± 0.03 & 0.02 ± 0.01 & - & 0.00 ± 0.02 & 0.00 ± 0.02 & 0.01 ± 0.00 & - \\
Llama:8b & - & - & 0.10 ± 0.02 & - & - & - & - & - & - & - \\
Llama:70b & - & - & 0.08 ± 0.01 & - & - & - & - & - & - & - \\
Llama-405b & 0.09 ± 0.03 & 0.20 ± 0.04 & 0.47 ± 0.03 & 0.47 ± 0.05 & 0.19 ± 0.04 & - & 0.28 ± 0.04 & 0.34 ± 0.05 & 0.32 ± 0.04 & - \\
Mistral-Nemo & - & - & 0.03 ± 0.01 & - & - & - & - & - & - & -\\
\end{tabular}
\end{adjustbox}
\caption{The success rate by model and dataset}
\label{tab:model_performance_success_rate}
\end{table}

\begin{table}[ht]
\centering
\renewcommand{\arraystretch}{1.2}
\begin{adjustbox}{width=1.5\textwidth}
\begin{tabular}{l|ccccc|ccccc}
\textbf{} & \multicolumn{5}{c|}{\textbf{Wikipedia}}& \multicolumn{5}{c}{\textbf{OpenAlex}} \\
\textbf{Model} & \textbf{Bengali} & \textbf{Chinese} & \textbf{English} & \textbf{French} & \textbf{Russian} & \textbf{Arts} & \textbf{Biology} & \textbf{Medicine} & \textbf{Physics} & \textbf{Social Sciences} \\ \hline
TF-IDF & 25.11 ± 0.95 & \textbf{25.21 ± 1.45} & 32.88 ± 1.13 & 30.24 ± 1.02 & 36.51 ± 1.31 & 28.50 ± 3.08 & 37.43 ± 1.22 & 42.61 ± 1.23 & 45.28 ± 1.44 & 41.14 ± 2.14 \\ \hdashline
AllMPNetBase & x & 43.25 ± 2.15 & 32.97 ± 0.36 & 33.42 ± 1.17 & 43.20 ± 1.10 & \textbf{23.82 ± 0.55} & \textbf{25.79 ± 0.25} & \textbf{27.70 ± 0.26} & 31.58 ± 0.28 & \textbf{24.13 ± 0.35} \\
Distiluse-Base-v1 & x & 31.47 ± 0.42 & 36.44 ± 0.47 & 36.62 ± 0.45 & 38.73 ± 0.55 & x & 44.57 ± 0.47 & 47.93 ± 0.50 & 47.74 ± 0.51 & 38.02 ± 0.58 \\
Distiluse-Base-v2 & x & 34.23 ± 0.45 & 36.38 ± 0.46 & 35.67 ± 0.42 & 37.55 ± 0.48 & x & 43.87 ± 0.46 & 47.90 ± 0.50 & 48.47 ± 0.51 & 37.73 ± 0.57 \\
E5-Large & 27.54 ± 0.33 & 29.48 ± 0.32 & 35.78 ± 0.35 & 32.17 ± 0.32 & 34.17 ± 0.35 & 30.04 ± 0.81 & 31.57 ± 0.28 & 36.55 ± 0.31 & 39.25 ± 0.34 & 30.73 ± 0.41 \\
E5-Small & 27.67 ± 0.32 & 28.50 ± 0.31 & 35.53 ± 0.35 & 32.38 ± 0.33 & 32.99 ± 0.35 & x & 35.60 ± 0.31 & 39.77 ± 0.34 & 43.12 ± 0.39 & 34.26 ± 0.47 \\
MiniLM-L6 & x & 41.88 ± 0.81 & 33.91 ± 0.37 & 34.64 ± 0.40 & 45.75 ± 1.29 & 27.38 ± 0.68 & 29.62 ± 0.29 & 32.40 ± 0.29 & 35.84 ± 0.31 & 27.54 ± 0.39 \\
MiniLM-L12 & 36.47 ± 0.95 & 32.89 ± 0.49 & 38.87 ± 0.50 & 36.39 ± 0.46 & 36.22 ± 0.54 & x & 41.34 ± 0.41 & 44.59 ± 0.43 & 46.08 ± 0.47 & 34.24 ± 0.56 \\
Para-MPNet-Base & 33.34 ± 0.50 & 33.57 ± 0.50 & 38.66 ± 0.49 & 35.75 ± 0.43 & 37.03 ± 0.51 & x & 41.33 ± 0.42 & 42.89 ± 0.44 & 46.52 ± 0.49 & 32.73 ± 0.54 \\
STSB-XLM-R & 35.66 ± 0.55 & 34.53 ± 0.57 & 40.39 ± 0.50 & 37.77 ± 0.45 & 38.26 ± 0.55 & x & 48.62 ± 0.66 & 52.62 ± 0.79 & 52.52 ± 0.72 & 40.13 ± 0.73 \\ 
Albert-Small & - & - & 39.74 ± 0.52 & - & - & x & 38.80 ± 0.36 & 42.22 ± 0.39 & 46.12 ± 0.43 & 35.57 ± 0.55 \\
DistilRoberta & - & - & 34.92 ± 0.39 & - & - & 26.81 ± 0.69 & 29.41 ± 0.28 & 32.35 ± 0.31 & 34.72 ± 0.31 & 26.37 ± 0.40 \\
Glove & - & - & 37.25 ± 0.52 & - & - & x & 44.04 ± 0.50 & 50.68 ± 0.65 & 48.58 ± 0.57 & 40.27 ± 0.71 \\
Roberta & - & - & 35.40 ± 0.38 & - & - & 25.35 ± 0.60 & 27.23 ± 0.26 & 30.12 ± 0.27 & 33.64 ± 0.30 & \textbf{23.98 ± 0.36} \\\hdashline
GPT-4o mini & \textbf{19.38 ± 2.18} & \textbf{27.36 ± 3.08} & \textbf{26.47 ± 3.44} & 29.65 ± 3.02 & \textbf{25.26 ± 4.16} & - & 44.84 ± 3.54 & 50.55 ± 4.00 & 42.43 ± 3.56 & - \\
GPT-4o & - & - & \textbf{21.55 ± 2.22} & \textbf{18.93 ± 2.08} & - & - & - & - & \textbf{27.05 ± 2.17} & - \\ \hdashline
DS-R1:70B (R) & 21.87 ± 3.05 & 19.84 ± 3.11 & 21.37 ± 1.71 & 26.09 ± 3.20 & 21.50 ± 2.83 & - & 32.09 ± 3.28 & 36.76 ± 3.46 & 37.78 ± 3.56 & - \\ \hdashline
Gemma2:27B & 27.29 ± 4.18 & 45.00 ± 7.15 & 39.54 ± 2.88 & 52.08 ± 5.49 & 43.81 ± 6.08 & - & 45.50 ± 6.45 & 67.63 ± 5.53 & 54.85 ± 7.47 & - \\
Llama:1b & - & - & x & - & - & - & - & - & - & - \\
Llama:3b & 31.78 ± 5.22 & x & 35.21 ± 3.18 & x & 40.67 ± 23.73 & - & x & x & x & - \\
Llama:8b & - & - & 36.21 ± 3.94 & - & - & - & - & - & - & - \\
Llama:70b & - & - & 41.55 ± 3.12 & - & - & - & - & - & - & - \\
Llama:405b & 29.10 ± 9.76 & 36.10 ± 5.74 & 27.61 ± 2.29 & 26.74 ± 3.49 & 36.40 ± 5.88 & - & 57.46 ± 4.80 & 53.06 ± 3.48 & 48.00 ± 3.74 & - \\
Mistral-Nemo & - & - & x & - & - & - & - & - & - & - \\
\end{tabular}
\end{adjustbox}
\caption{The mean number of steps above optimal for successful games by model and dataset. When the success rate is less than or equal to 0.05, the values are not shown and are marked with an `x'.}
\label{tab:model_performance_mean_steps}
\end{table}
\end{landscape}

\clearpage

\section{Distribution Fitting}\label{appendix_distribution_fitting}
\typeout{~~ Distribution Fitting ~~}

To understand the behaviour of the agents the distribution of the step count was studied.
The form which most closely matched the pattern seen is a negative binomial distribution. 
This can be thought of as a count of the number of failed attempts to do something with probability $p$ before it happens $n$ times.
If the total number of failures is given by $x$ then the probability function follows 
\begin{equation}
    \Pr(x) = \binom{x + n - 1}{x} (1 - p)^x p^n \quad .
\end{equation}

\begin{figure}[ht]
    \centering
    \includegraphics[width=0.5\linewidth]{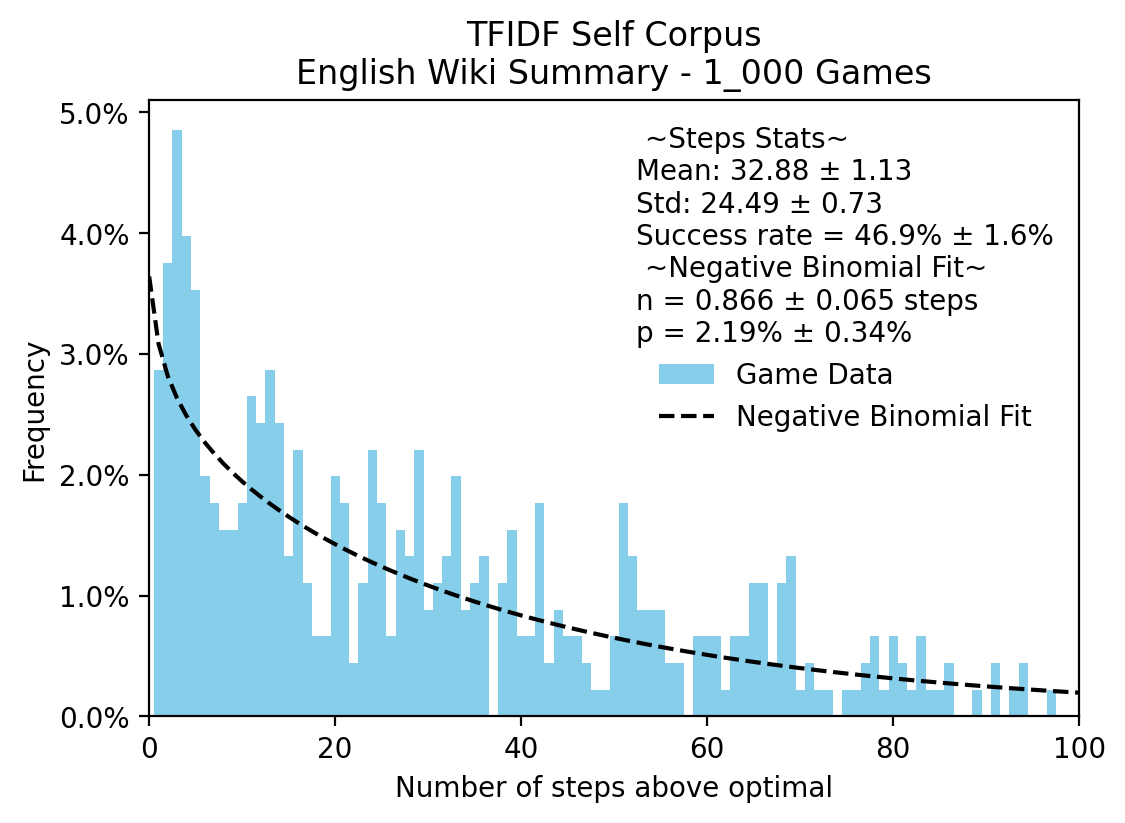}
    \caption{Histogram of number of steps above optimal for the TF-IDF model on the English Wikipedia dataset}
    \label{fig:tfidf_en_wiki}
\end{figure}

One way to imagine the Wikipedia game is to think of a path that needs to be followed to the target article.
If each step along that path could happen with probability $p$ and if the misstep made no positive or negative progress, then the number of extra steps taken by the agents would follow the negative binomial distribution.
The optimum number of steps for each of the pairs of pages is known, so it is possible to plot these extra steps to see if the distribution is a good fit. 
\linkref{fig:tfidf_en_wiki}{Figure}{} through to \linkref{fig:qwq32b_en_wiki}{Figure}{} show histograms of the number of steps above optimal for eight models on the English Wikipedia dataset.

What is interesting is that while the negative binomial distribution seems to be a good fit for these models, the value of $n$ is strange.
For all of the models, the expected optimum from the fit seems to be statistically indistinguishable from 1.
This is interesting as it implies that there is one major step that seems to be dominating the performance of the agents.
This behaviour is also not seen when humans play games which have an $n$ value around 3 (when fitting to all games), much closer to the real optimum value (see \linkref{fig:humans_hist}{Figure}{}).
This suggests fundamental differences between how humans and language models deal with Wikipedia games and speaks against tempting anthropomorphising comparisons seen between the success rate of humans and GPT-4o.

For the human games, we repeated the fit, treating games more than 10 steps above optimal as unknown, since the 10-minute time limit is most likely to cut short longer games (\linkref{fig:human_nb_fits}{Figure}{}).
With $n$ free, the fit yields $n = 4.45 \pm 2.15$, while fixing $n = 1$ (a geometric distribution) increases the negative log-likelihood by 6.58 with one fewer parameter (likelihood-ratio test, $p \approx 3 \times 10^{-4}$) and implies an unphysical eventual success rate of 110\%.
Constraining that rate to at most 100\% could only worsen the $n = 1$ fit, so the test is conservative.

\begin{figure}[ht]
    \centering
    \includegraphics[width=0.5\linewidth]{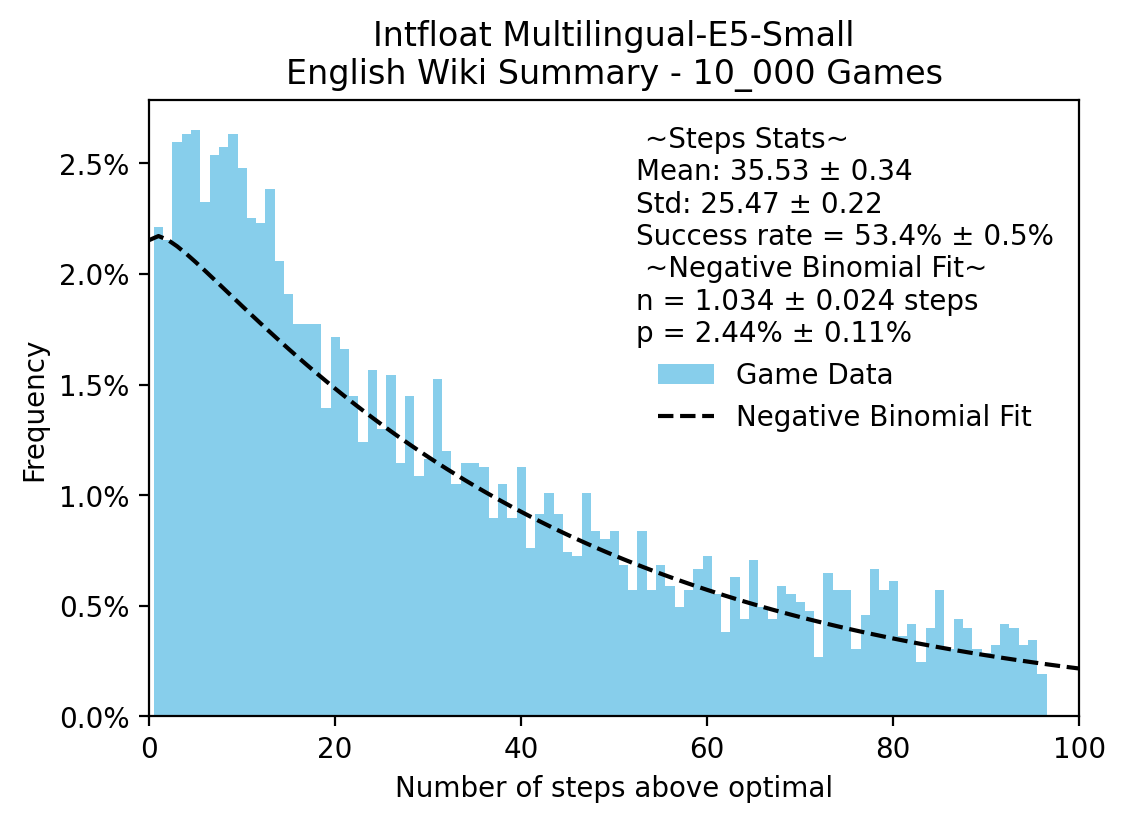}
    \caption{Histogram of number of steps above optimal for the E5-Small model on the English Wikipedia dataset}
    \label{fig:e5_small_en_wiki}
\end{figure}

\begin{figure}[ht]
    \centering
    \includegraphics[width=0.5\linewidth]{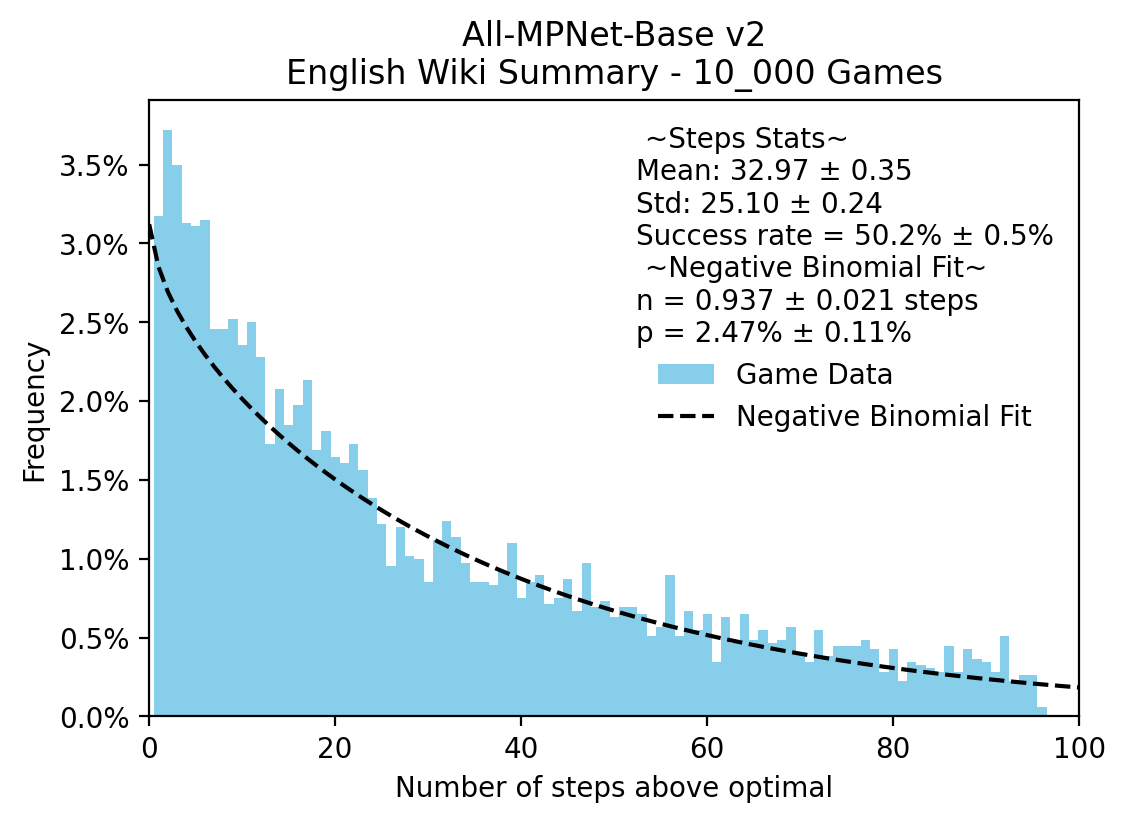}
    \caption{Histogram of number of steps above optimal for the AllMPNetBase model on the English Wikipedia dataset}
    \label{fig:mpnet_en_wiki}
\end{figure}

\begin{figure}[ht]
    \centering
    \includegraphics[width=0.5\linewidth]{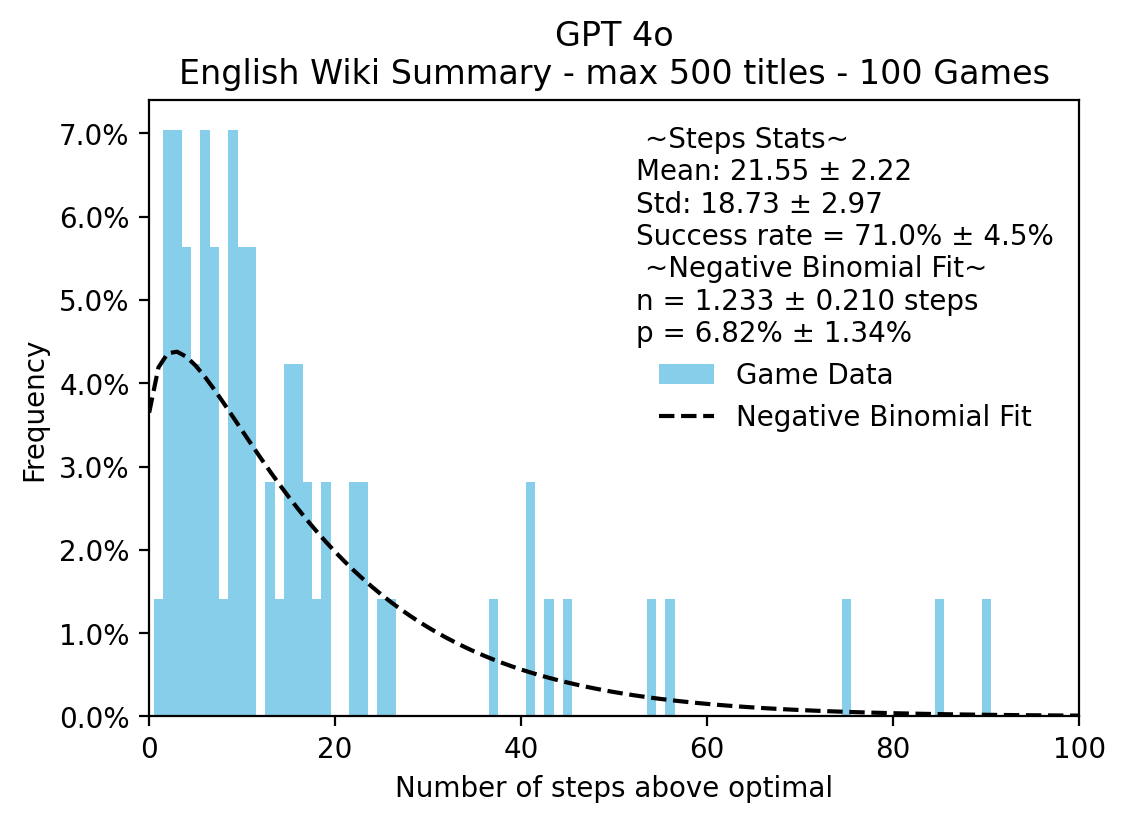}
    \caption{Histogram of number of steps above optimal for the GPT-4o model on the English Wikipedia dataset}
    \label{fig:gpt4o_en_wiki}
\end{figure}

\begin{figure}[ht]
    \centering
    \includegraphics[width=0.5\linewidth]{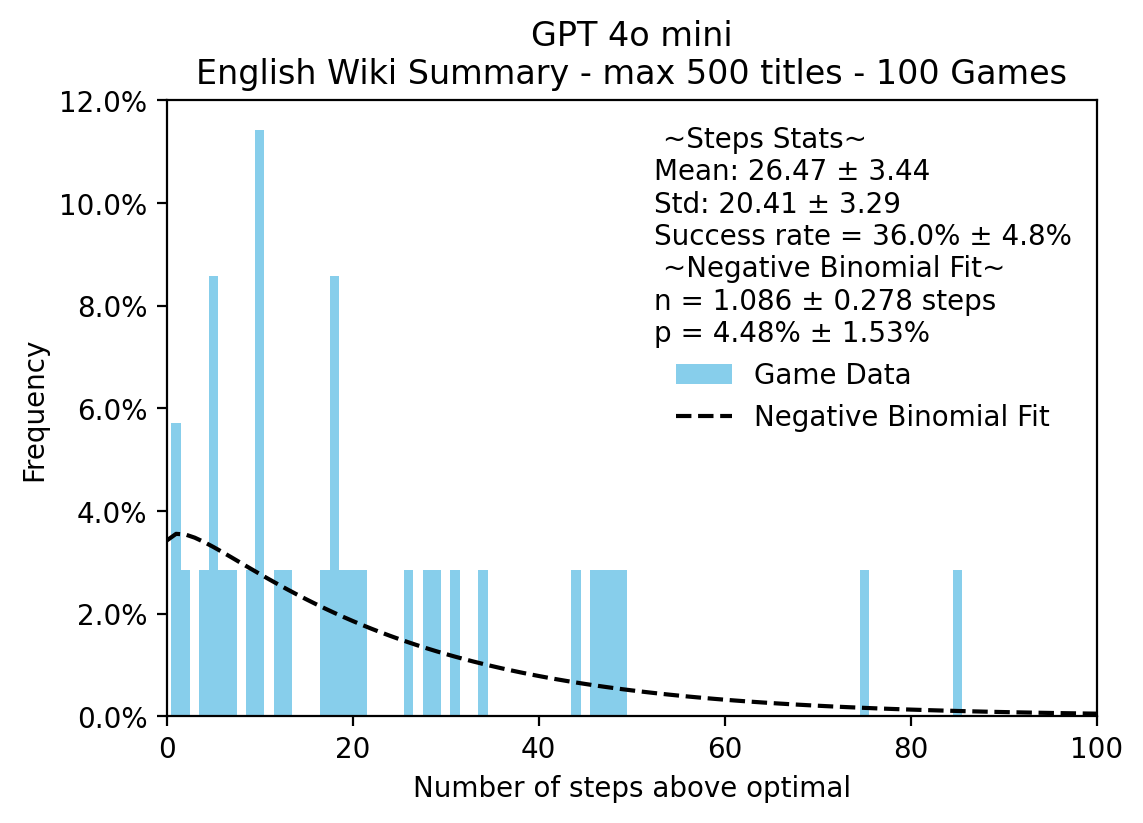}
    \caption{Histogram of number of steps above optimal for the GPT-4o mini model on the English Wikipedia dataset}
    \label{fig:gpt4o_mini_en_wiki}
\end{figure}

\begin{figure}[ht]
    \centering
    \includegraphics[width=0.5\linewidth]{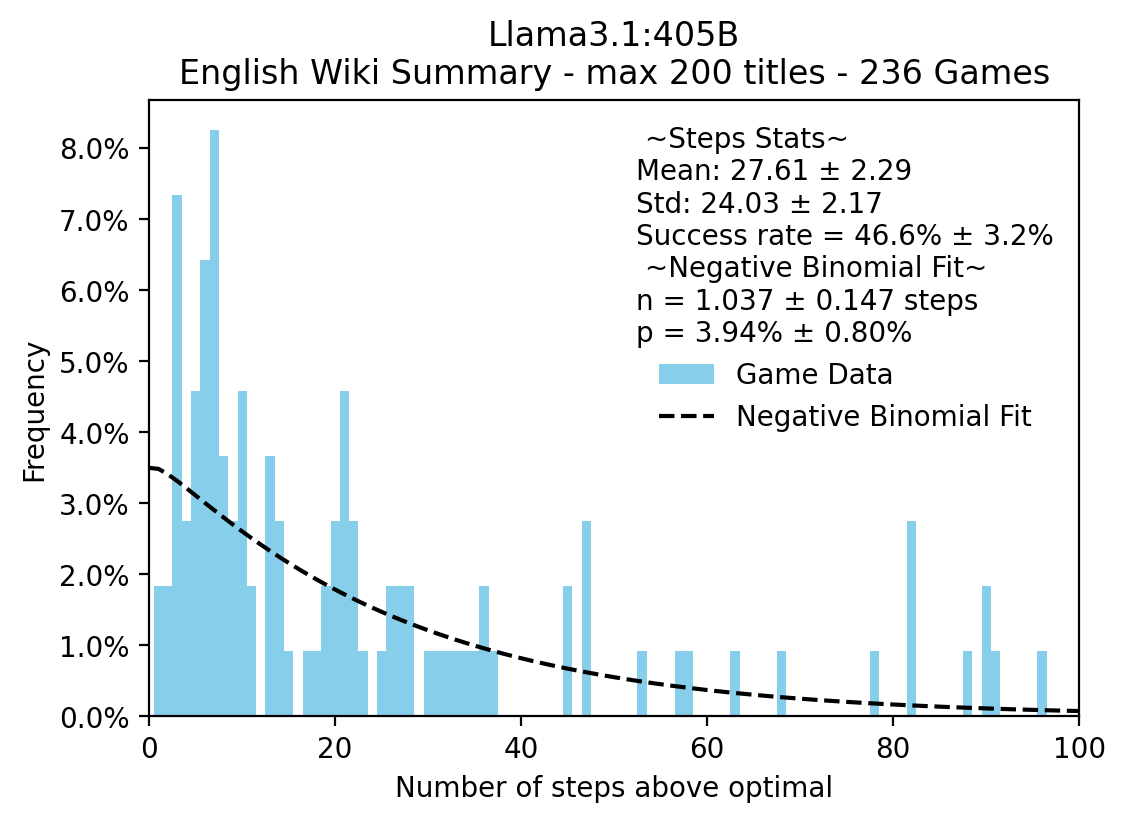}
    \caption{Histogram of number of steps above optimal for the Llama:405b model on the English Wikipedia dataset}
    \label{fig:llama405b_en_wiki}
\end{figure}

\begin{figure}[ht]
    \centering
    \includegraphics[width=0.5\linewidth]{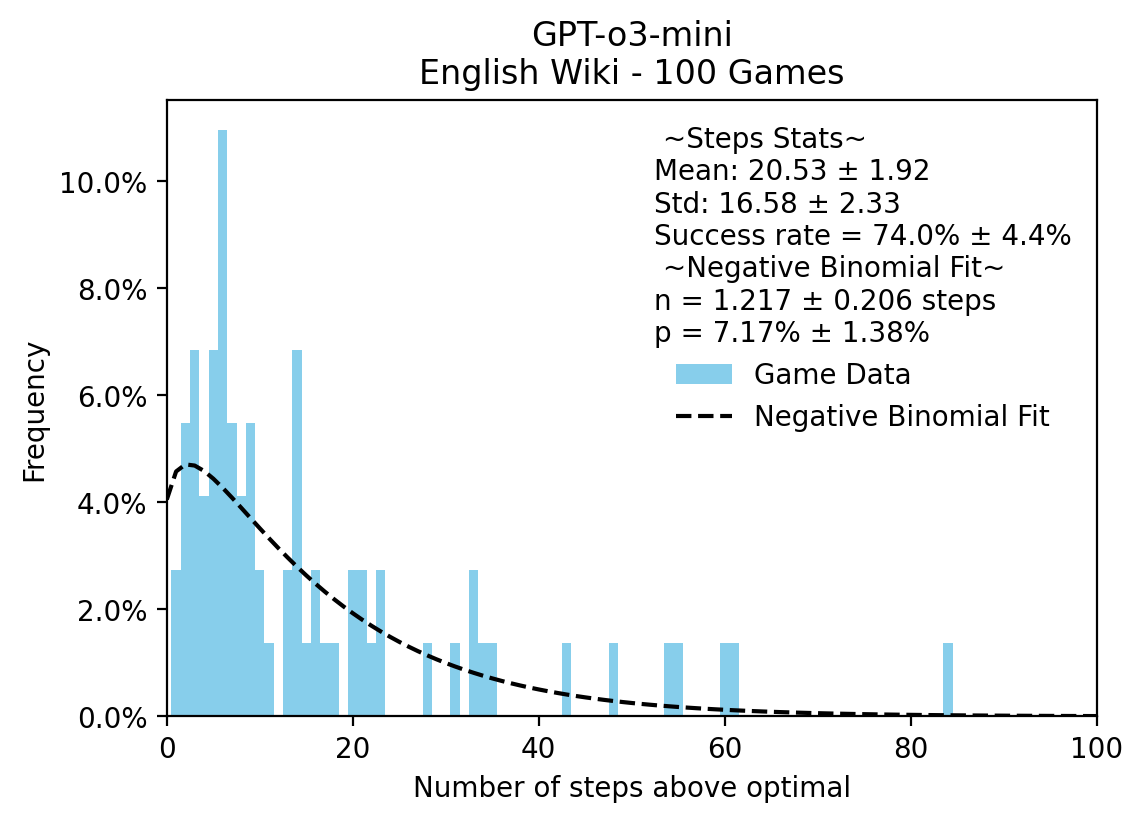}
    \caption{Histogram of number of steps above optimal for the GPT-o3 Mini Reasoning (R) model on the English Wikipedia dataset}
    \label{fig:gpto3mini_en_wiki}
\end{figure}

\begin{figure}[ht]
    \centering
    \includegraphics[width=0.5\linewidth]{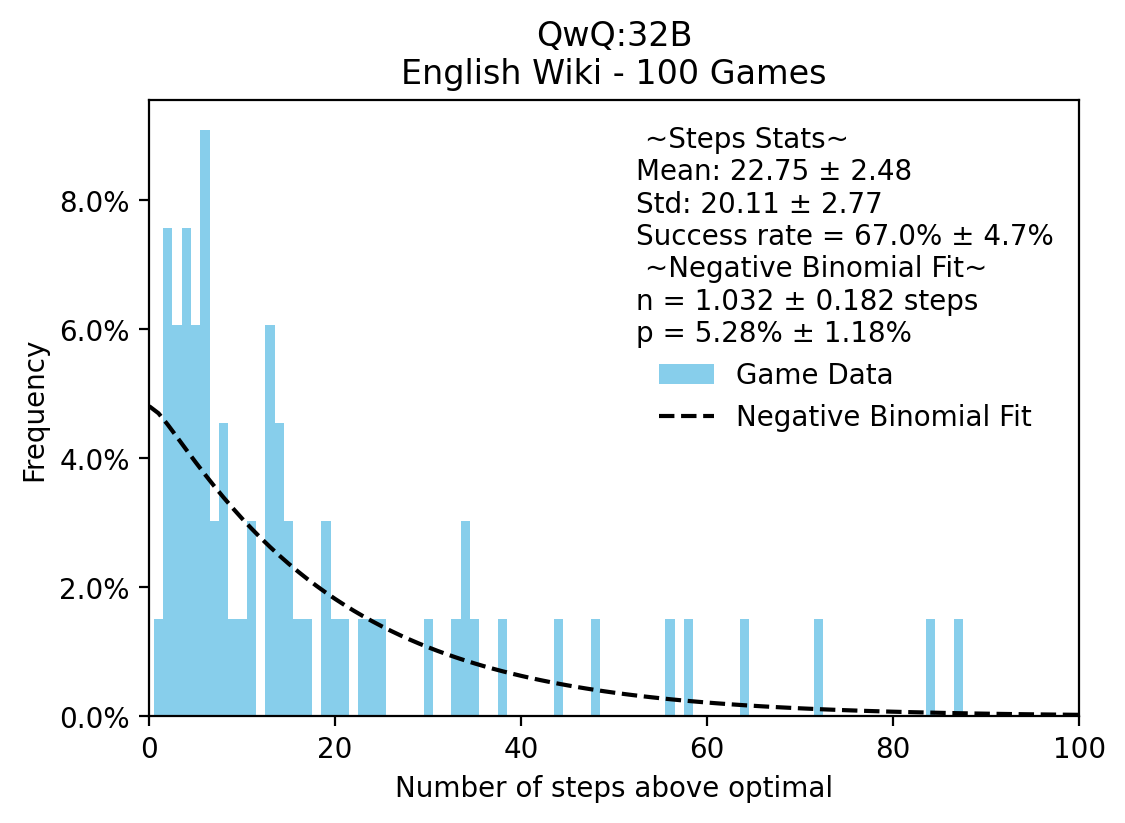}
    \caption{Histogram of number of steps above optimal for the QwQ:32B Reasoning (R) model on the English Wikipedia dataset}
    \label{fig:qwq32b_en_wiki}
\end{figure}

\begin{figure}[ht]
    \centering
    \includegraphics[width=0.5\linewidth]{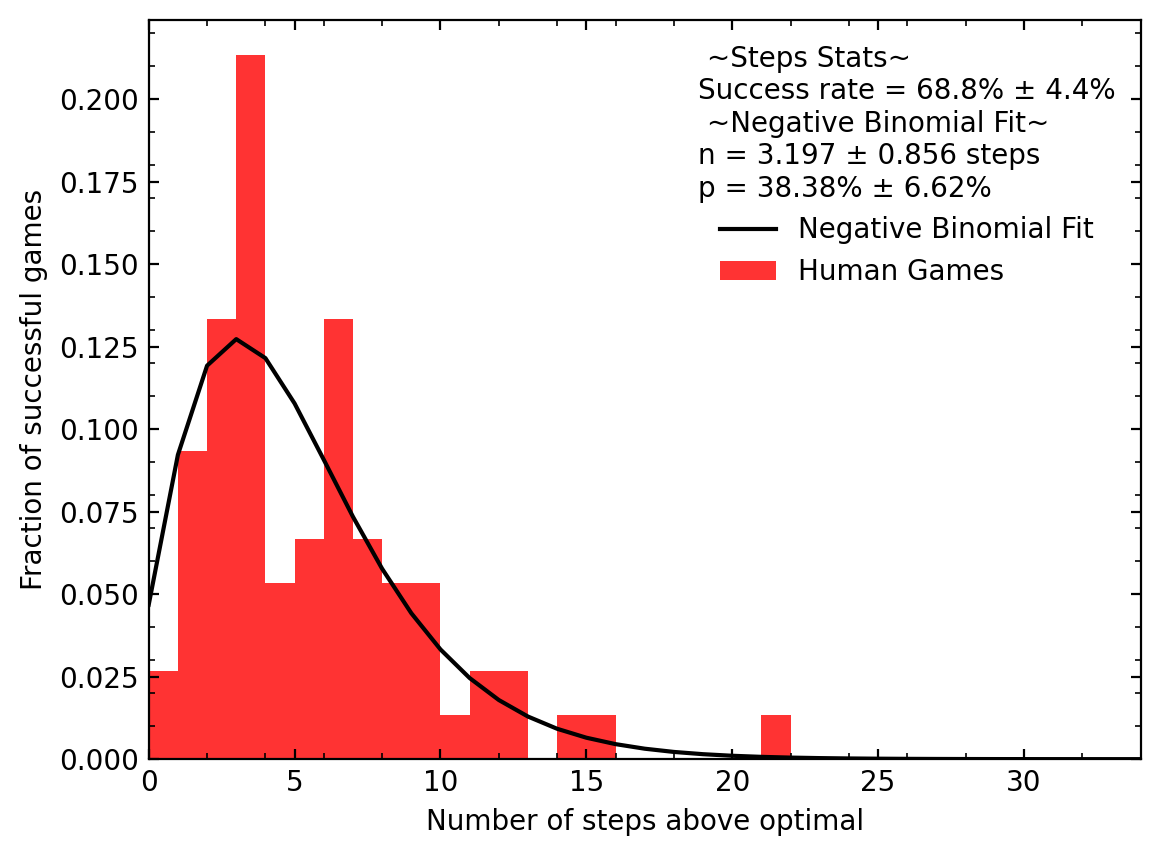}
    \caption{Histogram of number of steps above optimal for the successful games that the humans played.}
    \label{fig:humans_hist}
\end{figure}

\begin{figure}[ht]
    \centering
    \includegraphics[width=0.7\linewidth]{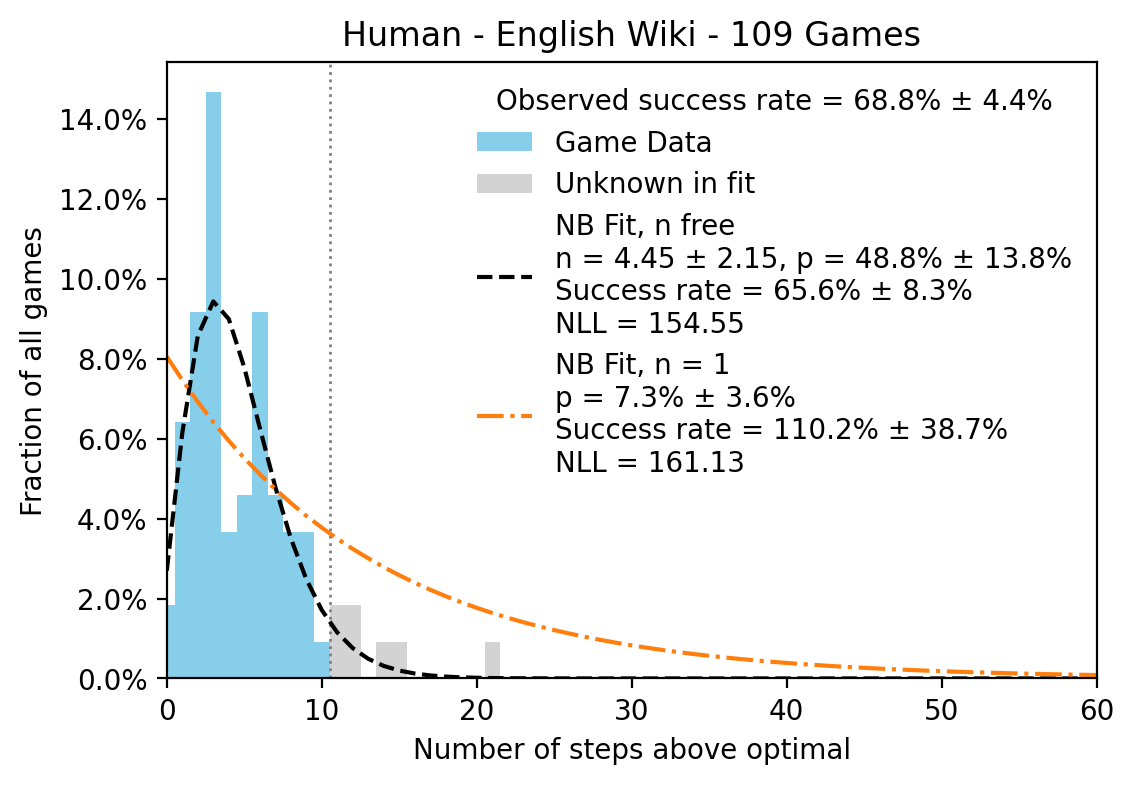}
    \caption{
    Negative binomial fits to the human games on English Wikipedia, with $n$ free (dashed) and with $n$ fixed at 1 (dash-dot).
    Games more than 10 steps above optimal (grey bars) are treated as unknown to reduce the effect of the time limit.
    }
    \label{fig:human_nb_fits}
\end{figure}

\clearpage

\section{Model Performance - Link Titles}\label{appendix_title_performance}
\typeout{~~ Title Performance ~~}

\linkref{tab:model_performance_success_rate_titles}{Table}{} shows the success rate on English Wikipedia and OpenAlex Physics dataset for a sample of sentence embedding models. The table shows the success rate in two scenarios. The first is when the summary for each link in an article is provided and compared to the summary of the target article (these are the same values as shown previously in \linkref{tab:model_performance_success_rate}{Table}{}). The second is the success rate when only the titles of the linked articles are provided and are compared to the summary of the target article. The use of the titles only makes the information given to the model comparable to that given to the LLM.

For English Wikipedia, the results show a significant drop in success rate for all the embedding models. For the OpenAlex Physics dataset, a significant drop in success rate also occurs for all models other than E5-Large and E5-Small. The E5-Large model has a more modest drop of 0.02 and the E5-Small model achieves an increase in success rate when only the titles are used. 

\begin{table}[ht]
\centering
\renewcommand{\arraystretch}{1.2}
\begin{adjustbox}{width=0.8\textwidth}
\begin{tabular}{l|cc|cc}
 & \multicolumn{2}{c|}{\textbf{Wikipedia:   English}} & \multicolumn{2}{c}{\textbf{OpenAlex:   Physics}} \\
 & \textbf{Summaries} & \textbf{Titles} & \textbf{Summaries} & \textbf{Titles} \\ \hline
AllMPNetBase & 0.50 ± 0.00 & 0.34 ± 0.05 & 0.63 ± 0.00 & 0.34 ± 0.00 \\
DistilRoberta & 0.44 ± 0.00 & 0.30 ± 0.05 & 0.58 ± 0.00 & 0.27 ± 0.00 \\
E5-Large & 0.53 ± 0.00 & 0.29 ± 0.05 & 0.52 ± 0.00 & 0.50 ± 0.00 \\
E5-Small & 0.53 ± 0.00 & 0.17 ± 0.04 & 0.41 ± 0.00 & 0.49 ± 0.00 \\
MiniLM-L6 & 0.47 ± 0.00 & 0.35 ± 0.05 & 0.57 ± 0.00 & 0.33 ± 0.00 \\
Roberta & 0.47 ± 0.00 & 0.42 ± 0.05 & 0.61 ± 0.00 & 0.26 ± 0.00
\end{tabular}
\end{adjustbox}
\caption{The success rate on English Wikipedia and OpenAlex Physics dataset for some sentence embedding models when only the titles of the link articles are used compared to when the summaries of the link articles are used.}
\label{tab:model_performance_success_rate_titles}
\end{table}

\clearpage

\section{Failure Reasons}\label{appendix_failure_reasons}
\typeout{~~ Failure Reasons ~~}

The games played with the sentence embedding and TF-IDF agents can have three possible outcomes:
\begin{itemize}
    \item \textbf{Success}: The agent navigates to the target article.
    \item \textbf{Failure - Step Limit Reached}: The agent fails to reach the target article within a set number of steps (configured to 100 in our games).
    \item \textbf{Failure - Dead End Reached}: The agent fails to reach the target article as it either reaches an article without any outward links. Since most agents are deterministic, articles that had already been visited were excluded from the options to avoid the agents becoming trapped in loops, but this means that if an agent reaches an article with no unseen articles, then this is a dead-end.
\end{itemize}

There are three more possible failure reasons when playing the game with the prompt agents:
\begin{itemize}
    \item \textbf{Failure - Hallucination}: The LLM hallucinates in its response by either not providing the response in valid JSON format, or by making up or not providing an article title to navigate to next.
    \item \textbf{Failure - Bad Request}: The API rejects the request, which might be what is suspected to be explicit or harmful content in the prompt.
    \item \textbf{Failure - No Response}: This may occur due to a timeout by the API.
\end{itemize}

On top of these was also one extra failure mode when using the reasoning prompt:
\begin{itemize}
    \item \textbf{Failure - No Answer}: The result the model produced did not contain an answer block.
\end{itemize}

\begin{figure}[!ht]
    \centering
    \includegraphics[width=0.85\linewidth]{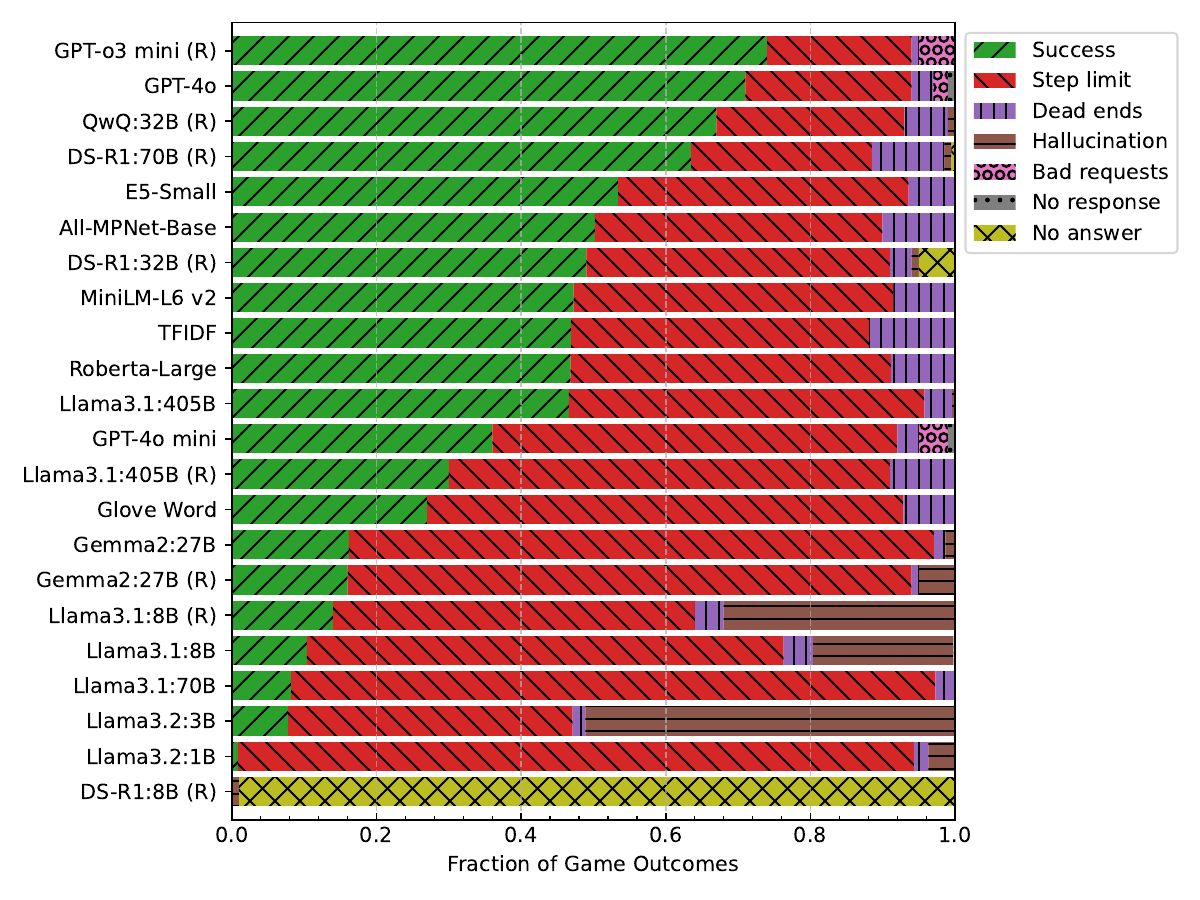}
    \caption{\protect\input{sup_material/fig_failure/acaption_failure_reasons_enwiki}}
    \label{afig:failure_enwiki}
\end{figure}

The outcomes of the games for the English Wikipedia dataset and a selection of models are shown in \linkref{afig:failure_enwiki}{Figure}{}. This shows that reaching the step limit is the most common failure reason for the majority of the models. One exception is Llama:3b, which has a high proportion of games failing in hallucinations. It is notable that the proportion of games failing due to hallucinations is significantly higher for Llama:3b than for Llama:1b. This could be a result of the Llama:1b model selecting article titles from the given list randomly, without any reasoning about how to play the game. The GPT models are the only ones that fail due to bad API requests owing to the safeguards that are in place on Azure's OpenAI service.

\begin{figure}[!ht]
    \centering
    \includegraphics[width=0.85\linewidth]{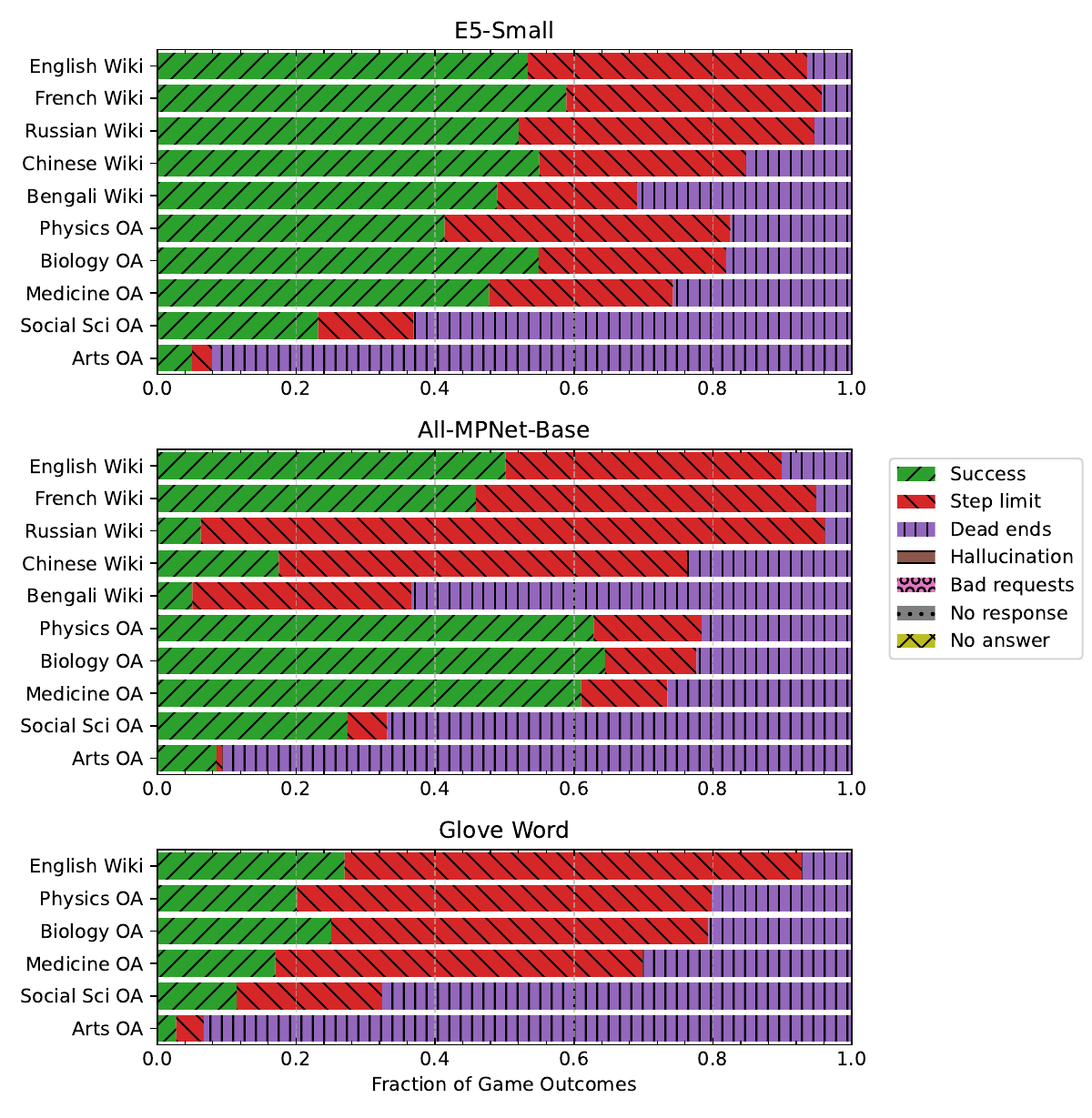}
    \caption{\protect\input{sup_material/fig_failure/acaption_failure_by_topic_embed}}
    \label{afig:failure_topic}
\end{figure}

The outcomes for games using sentence embedding models across all the datasets are shown in \linkref{afig:failure_topic}{Figure}{}. These show the high proportion of games fail for the Arts and Social Sciences datasets because they reach a dead end in the graph.
This is not a surprising result given the poor connectivity of the graphs, as shown in \linkref{appendix_datasets}{Appendix}{}. 

\linkref{afig:failure_topic_llm}{Figure}{} shows the outcomes by dataset for a selection of LLMs. These show the high proportion of games that fail due to hallucinations for Llama:3b; although hallucinations occur across all datasets, the proportion of hallucinations on the OpenAlex datasets is particularly high.

\begin{figure}[!ht]
    \centering
    \includegraphics[width=0.85\linewidth]{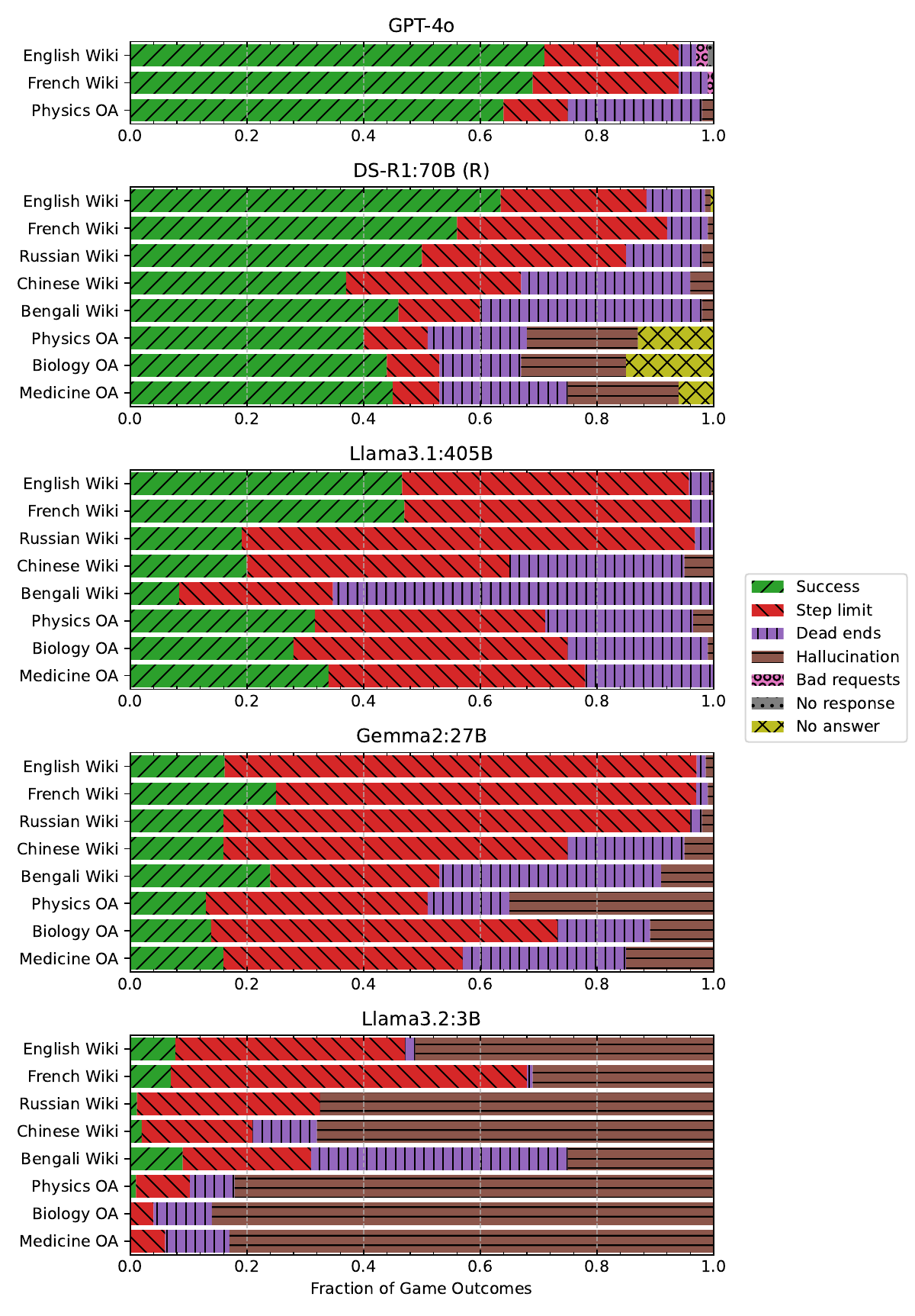}
    \caption{\protect\input{sup_material/fig_failure/acaption_failure_by_topic_llm}}
    \label{afig:failure_topic_llm}
\end{figure}

\clearpage

\section{TF-IDF Performance by Corpus}\label{appendix_tfidf_corpus}
\typeout{~~ TF-IDF Performance by Corpus ~~}

\linkref{tab:tfidf_corpus}{Table}{} shows the success rate for the TF-IDF model on the OpenAlex datasets in two scenarios. In the first scenario, a sample of 100,000 records from the target dataset is used to create the corpus for the TF-IDF model. In the second scenario, a sample of 100,000 records from the English Wikipedia dataset is used to create the corpus. 

The results show a drop in the success rate for the Biology, Medicine and Physics datasets when the English Wikipedia dataset is used for the corpus. However, the success rate for Social Sciences increases when the English Wikipedia dataset is used. This suggests that there is a bigger overlap between the vocabulary of the Social Sciences dataset with English Wikipedia than for Biology, Medicine and Physics.

\begin{table}[ht]
\centering
\renewcommand{\arraystretch}{1.2}
\begin{adjustbox}{width=1\textwidth}
\begin{tabular}{l|ccccc}
\textbf{} &  \multicolumn{5}{c}{\textbf{OpenAlex}} \\
\textbf{Corpus} & \textbf{Arts and Humanities} & \textbf{Biology} & \textbf{Medicine} & \textbf{Physics} & \textbf{Social Sciences} \\ \hline
Target Dataset & 0.04 ± 0.01 & 0.43 ± 0.02 & 0.34 ± 0.02 & 0.33 ± 0.01 & 0.15 ± 0.01 \\
English Wikipedia & 0.04 ± 0.01 & 0.14 ± 0.01 & 0.27 ± 0.01 & 0.25 ± 0.01 & 0.35 ± 0.02 \\
\end{tabular}
\end{adjustbox}
\caption{The success rate of TF-IDF by corpus and dataset}
\label{tab:tfidf_corpus}
\end{table}
\clearpage

\section{Example Routes}\label{appendix_routes}
\typeout{~~ Example Routes ~~}

This section shows some examples of routes taken by some of the models. These demonstrate how they can behave differently and some circumstances where one model might succeed and another fails. 

\subsection{`Afloqualone' to `Gaurotes pictiventris'}
\linkref{lst_gpt_success}{Listing}{} shows the route taken (with reasoning) by GPT-4o from the article titled \href{https://en.wikipedia.org/wiki/Afloqualone}{Afloqualone} to \href{https://en.wikipedia.org/wiki/Gaurotes_pictiventris}{Gaurotes pictiventris} (a species of beetle) from the English Wikipedia graph. In the example shown, GPT-4o was able to navigate the target article in 13 steps. None of the sentence embedding model were able to navigate to the target article within 100 steps. At the time of writing, the summary for the target article contains limited relevant information:

\begin{quote}
Gaurotes pictiventris is a species of beetle in the family Cerambycidae. It was described by Pesarini and Sabbadini in 1997.Bezark, Larry G. A Photographic Catalog of the Cerambycidae of the World. Retrieved on 22 May 2012.  References  Category:Lepturinae Category:Beetles described in 1997
\end{quote}

As well has having limited relevant semantic information in the article summary, the article has only two links into the article which will make it more difficult to navigate to.

\begin{lstlisting}[breaklines,basicstyle=\ttfamily,caption=Example of a route taken by GPT-4o when navigating to `Gaurotes pictiventris' with the reasoning given by the model at each step,label=lst_gpt_success]
0 : Afloqualone - Start page
1 : Dermatitis - Dermatitis is a condition that can be caused by various factors, including insect bites. This could potentially lead to a link about insects or beetles, which might help us navigate towards the target page about the beetle species Gaurotes pictiventris.
2 : Bacteria - Since Gaurotes pictiventris is a species of beetle, it belongs to the biological classification of organisms. Bacteria are also a type of organism, and navigating to a page about bacteria might provide links to other biological classifications, including insects and beetles, which could help us get closer to the target page.
3 : Bacteriology - Bacteriology is the study of bacteria, which includes the classification and description of bacterial species. Since Gaurotes pictiventris is a species of beetle, it is likely that the page on Bacteriology will have links to pages related to taxonomy and species classification, which can help navigate towards the target page about a specific beetle species.
4 : Taxonomy (biology) - The target page `Gaurotes pictiventris' is about a species of beetle, which falls under the study of taxonomy in biology. Taxonomy deals with the classification and naming of organisms, which is relevant to finding information about a specific species of beetle.
5 : Family (biology) - Since Gaurotes pictiventris is a species of beetle in the family Cerambycidae, navigating to the `Family (biology)' page will provide a direct link to the concept of biological families, which is closely related to the target page. From there, you can likely find links to specific families of beetles, including Cerambycidae.
6 : Subfamily - Gaurotes pictiventris is a species of beetle in the family Cerambycidae. To navigate to the target page, it is logical to start with a more specific taxonomic rank related to beetles. `Subfamily' is a relevant choice as it is a taxonomic rank within the family, which can lead us closer to the specific species.
7 : Zoology - Gaurotes pictiventris is a species of beetle, which falls under the study of animals. Zoology is the branch of biology that deals with the animal kingdom, so it is the most relevant title to navigate towards the target page.
8 : Entomology - Entomology is the scientific study of insects, and since Gaurotes pictiventris is a species of beetle, which is an insect, this link is likely to lead you closer to information about beetles and potentially to the family Cerambycidae.
9 : Beetle - Gaurotes pictiventris is a species of beetle in the family Cerambycidae. Selecting `Beetle' will likely lead to more specific information about beetles, which can help narrow down to the specific species.
10 : Longhorn beetle - Gaurotes pictiventris is a species of beetle in the family Cerambycidae, which is commonly known as longhorn beetles. Navigating to the `Longhorn beetle' page will likely provide links or information related to the Cerambycidae family, helping you get closer to the target page.
11 : Lepturinae - Gaurotes pictiventris belongs to the family Cerambycidae, and Lepturinae is a subfamily within Cerambycidae. This link is likely to lead you closer to the target page as it is more specific to the classification of the beetle.
12 : Gaurotes - Gaurotes pictiventris is a species of beetle in the family Cerambycidae, and `Gaurotes' is the genus to which this species belongs. Navigating to the genus page will likely provide relevant links or information to help you reach the specific species page.
13 : Gaurotes pictiventris - The title `Gaurotes pictiventris' directly matches the target page you are trying to navigate to. This is the exact species of beetle described in the summary you provided.
\end{lstlisting}

\subsection{`Magic star' to `Choir of St George's Chapel, Windsor Castle'}

The TF-IDF, E5-Small and GPT-4o models were all capable of navigating from \href{https://en.wikipedia.org/wiki/Magic_star}{Magic star} to \href{https://en.wikipedia.org/wiki/Choir_of_St_George%27s_Chapel,_Windsor_Castle}{Choir of St George's Chapel, Windsor Castle}. The summary of this target article is longer than the previous example:

\begin{quote}
right thumb  The Quire of St George's Chapel, by Charles Wild, from W.H. Pyne's Royal Residences, 1818. The Choir of St George's Chapel at Windsor Castle exists to sing services in St George's Chapel at Windsor Castle.  It has been in existence since 1348 and, with the exception of the Commonwealth period (1649–1660), has sung services in the Chapel continuously ever since.   The choir today  The choir comprises up to 23 choristers (the youngest of whom are probationers or training choristers) and 12 professional Lay Clerks, who sing alto, tenor and bass. The Choir sings at Evensong each day (except Wednesdays) and at Mattins and Eucharist on Sunday mornings during the three terms of the College year. The choristers are educated at St George's School, Windsor Castle which is situated in the Castle grounds. The Lay Clerks live in the Horseshoe Cloister and on Denton's Commons.  The choir sings regularly in the presence of the King and other members of the Royal Family.  Concerts are als
\end{quote}

There is some text in this summary that is irrelevant to the article and is left over from the markup (such as `right thumb' at the beginning of the article). The summary text also stops mid-sentence, when a limit on the text length has been reached. Both features are common across the Wikipedia articles. 

\linkref{lst_magic_star_tfidf}{Listing}{} shows the route taken by the TF-IDF model, which only needs 12 steps to reach the target article (the minimum possible number of steps is five). The cosine similarity values between the TF-IDF vectors of the linked article and the target article are also shown to two decimal places. 

\linkref{lst_magic_star_embedding}{Listing}{} shows the route taken by the E5-Small embedding model. This takes 41 steps to reach the target. 

\linkref{lst_magic_star_gpt}{Listing}{} shows the route taken by the GPT-4o model with the reasoning given by the model at each step. GPT-4o is able to reach the target article in ten steps; however, the model failed to pick the direct link to the target when it appeared in the article at step eight, so it is still not immune to errors. The model is able to reason that stepping to a general article like `Religion' is likely to lead to relevant links. There may be parallels in the choice of this step with how humans play the game, where participants often navigate to a general article likely to have many links, before narrowing down the search to the specifics of the target article. \linkref{tab:compare_routes}{Table}{} lists the article names in the route taken by GPT-4o and shows the cosine similarity between those article summaries and the target article summary for the TF-IDF and E5-Small models. This shows that the article titled `Religion' is not considered similar to the target by either of these models, making both less likely to select the article.

In contrast to GPT-4o, \linkref{lst_magic_star_llama3b}{Listing}{} shows the route taken by Llama:3b, which was not able to navigate to the target article. The output shows inaccurate reasoning from the second step where the reasoning states that the word `dodecahedron' is contained on the target article: it is not in the provided summary or anywhere on the Wikipedia article (at the time of writing). The model carries on selecting valid links, with many hallucinations in the reasoning, until it is unable to select a valid link at step 22.

\begin{lstlisting}[breaklines,basicstyle=\ttfamily,caption=Example of a route taken by a TF-IDF model that used English Wikipedia to create a corpus when navigating to `Choir of St George's Chapel\, Windsor Castle' - the cosine similarity scores are shown between the current and target articles at each step,label=lst_magic_star_tfidf]
0 : Magic star : 0.01
1 : Magic hexagram : 0.00
2 : Hexagram : 0.01
3 : Ulster Banner : 0.03
4 : Dublin Castle : 0.12
5 : Chapel Royal, Dublin : 0.21
6 : Glenstal Abbey : 0.06
7 : Faith of Our Fathers (album) : 0.09
8 : Cor na nog : 0.27
9 : Choir : 0.32
10 : Choir of King's College, Cambridge : 0.47
11 : James Adcock : 0.44
12 : Choir of St George's Chapel, Windsor Castle : 1.00
\end{lstlisting}

\begin{lstlisting}[breaklines,basicstyle=\ttfamily,caption=Example of a route taken by E5-Small when navigating to `Choir of St George's Chapel\, Windsor Castle' - the cosine similarity scores are shown between the current and target articles at each step,label=lst_magic_star_embedding]
0 : Magic star : 0.73
1 : Magic constant : 0.75
2 : Magic series : 0.76
3 : Richard Schroeppel : 0.77
4 : Notices of the American Mathematical Society : 0.79
5 : English language : 0.76
6 : Court of Chancery : 0.82
7 : Master of the Rolls : 0.81
8 : Domus Conversorum : 0.82
9 : Public Record Office : 0.81
10 : Cyril Flower (historian) : 0.82
11 : St Edward's School, Oxford : 0.81
12 : Woodstock Road, Oxford : 0.82
13 : Observatory Street : 0.81
14 : Witney : 0.82
15 : George Wilkinson (architect) : 0.83
16 : Woodstock, Oxfordshire : 0.81
17 : Shipton-on-Cherwell : 0.81
18 : Bladon : 0.81
19 : Kidlington : 0.81
20 : River Cherwell : 0.80
21 : Bodicote : 0.82
22 : Adderbury : 0.81
23 : Ascot Racecourse : 0.82
24 : Windsor Castle Stakes : 0.82
25 : Harry Thomson Jones : 0.80
26 : St Leger Stakes : 0.81
27 : George Stanhope, 6th Earl of Chesterfield : 0.83
28 : George Stanhope, 7th Earl of Chesterfield : 0.82
29 : Cecil Weld-Forester, 1st Baron Forester : 0.82
30 : Orlando Weld-Forester, 4th Baron Forester : 0.82
31 : Baron Forester : 0.82
32 : Captain of the Honourable Corps of Gentlemen-at-Arms : 0.82
33 : Ivor Windsor-Clive, 2nd Earl of Plymouth : 0.82
34 : Robert Windsor-Clive (MP) : 0.82
35 : George Windsor-Clive (politician, born 1835) : 0.82
36 : George Windsor-Clive (politician, born 1878) : 0.82
37 : Parliament of the United Kingdom : 0.80
38 : Bishop of Winchester : 0.81
39 : William Day (bishop) : 0.83
40 : St George's Chapel, Windsor Castle : 0.86
41 : Choir of St George's Chapel, Windsor Castle : 1.00
\end{lstlisting}

\begin{lstlisting}[breaklines,basicstyle=\ttfamily,caption=Example of a route taken by GPT-4o when navigating to `Choir of St George's Chapel\, Windsor Castle' with the reasoning given by the model at each step,label=lst_magic_star_gpt]
0 : Magic star - Start page
1 : Star polygon - The Choir of St George's Chapel, Windsor Castle is related to a historical and architectural context.  `Star polygon' might lead to topics related to geometry and design, which could eventually connect to architectural elements of historical buildings like Windsor Castle.
2 : Religion - The Choir of St George's Chapel, Windsor Castle is closely associated with religious services and the Royal Family. `Religion' is the most relevant title from the list that could lead to information about religious institutions, services, and possibly specific chapels or choirs, including St George's Chapel.
3 : Religious music - The Choir of St George's Chapel, Windsor Castle is primarily known for its musical contributions to religious services. Starting from a page related to religious music could provide relevant links to choirs, chapels, or specific religious institutions, which can help navigate towards the target page.
4 : Anglican church music - The Choir of St George's Chapel, Windsor Castle is part of the Anglican tradition, and Anglican church music is directly related to the type of music performed by the choir. This link is likely to lead to more specific information about choirs within the Anglican Church, which can then guide you to the specific choir at St George's Chapel.
5 : Choir - The target page is about the Choir of St George's Chapel, Windsor Castle, and the title `Choir' is directly related to the concept of a choir, which is central to the target page's content. This should provide relevant links to navigate closer to the specific choir at Windsor Castle.
6 : Boys' choir - The Choir of St George's Chapel, Windsor Castle, includes choristers who are boys, making `Boys' choir' a relevant and direct link to explore more about the specific type of choir that includes young male singers.
7 : List of choir schools - The target page is about the Choir of St George's Chapel, Windsor Castle, which includes information about the choir's education at St George's School, Windsor Castle. `List of choir schools' is likely to have relevant links or information that can lead us closer to the target page.
8 : St George's School, Windsor Castle - The Choir of St George's Chapel, Windsor Castle is closely associated with St George's School, Windsor Castle, as the choristers are educated there. This connection makes it the most relevant link to navigate towards the target page.
9 : St George's Chapel, Windsor Castle - The Choir of St George's Chapel, Windsor Castle is directly related to St George's Chapel, Windsor Castle. This title will likely provide a direct link or relevant information leading to the target page.
10 : Choir of St George's Chapel, Windsor Castle - The title `Choir of St George's Chapel, Windsor Castle' directly matches the target page you are trying to navigate to. Selecting this title will take you directly to the target page.
\end{lstlisting}

\begin{table}[ht]
\centering
\renewcommand{\arraystretch}{1.5}
\begin{tabular}{lcc}
 & \multicolumn{2}{c}{\textbf{Cosine Similarity}}\\
\textbf{Article Title} & \textbf{E5-Small} & \textbf{TF-IDF} \\\hline
Magic star & 0.73 & 0.01 \\
Star polygon & 0.73 & 0.00\\
Religion & 0.73 & 0.00\\
Religious music & 0.77 & 0.00\\
Anglican church music & 0.80 & 0.11\\
Choir & 0.82 & 0.32\\
Boys' choir & 0.83 & 0.29\\
List of choir schools & 0.83 & 0.25\\
St George's School, Windsor Castle & 0.91 & 0.62\\
St George's Chapel, Windsor Castle & 0.86 & 0.39\\
Choir of St George's Chapel, Windsor Castle & 1.00 & 1.00\\
\hline
\end{tabular}
\caption{The cosine similarity between the named article and the target article of `Choir of St George's Chapel, Windsor Castle' for the E5-Small and TF-IDF models. The articles listed in the table are those taken by GPT-4o.}
\label{tab:compare_routes}
\end{table}

\clearpage

\begin{lstlisting}[breaklines,basicstyle=\ttfamily,caption=Example of a route taken by Llama:3b when navigating to `Choir of St George's Chapel\, Windsor Castle' with the reasoning given by the model at each step,label=lst_magic_star_llama3b]
0 : Magic star - Start page
1 : Schlafli symbol - The provided summary mentions `the Quire of St George's Chapel, by Charles Wild' which is likely related to the geometry or architecture of the chapel. Therefore, a title that relates to geometric shapes or symbols would be relevant.
2 : Dodecahedron - The target page `Choir of St George's Chapel, Windsor Castle' contains the phrase `dodecahedron', which is likely to be a reference to the Dodecahedron polytope. This seems like a plausible connection to make.
3 : Geometry - The target page `Choir of St George's Chapel, Windsor Castle' is related to the concept of geometry, which deals with the study of shapes and spatial relationships. The choir's structure and organization are likely to be discussed in a section on geometry.
4 : Euclidean geometry - The target page `Choir of St George's Chapel, Windsor Castle' is related to the history and tradition of singing in St George's Chapel at Windsor Castle. Euclidean geometry seems like a plausible starting point for navigation, as it deals with spatial relationships and measurement, which might be relevant to understanding the context of the choir's performances.
5 : Euclid - The summary mentions `Euclid's Elements' which is a strong hint towards the page about Euclid, a Greek mathematician. Additionally, there are multiple instances of `Euclid' in the list of titles, making it more likely that this is the correct page.
6 : Euclid's Elements - The target page `Choir of St George's Chapel, Windsor Castle' has a section titled `History', which mentions that the choir has sung services continuously since 1348. The next relevant page in the list is `Euclid's Elements', as Euclid was an ancient Greek mathematician who lived around 300 BCE and wrote one of the most influential mathematics books in history, making it a plausible connection to the historical context of the choir.
7 : Greek mathematics - The target page `Choir of St George's Chapel, Windsor Castle' has a section on the history of Greek mathematics, which is relevant to the topic. The title `Greek mathematics' appears in the list and seems like a plausible starting point.
8 : Hellenistic period - The Hellenistic period is mentioned in the provided summary as a historical context for the Choir of St George's Chapel, Windsor Castle. It seems like a relevant and connected topic to navigate to next.
9 : Hellenistic art - The target page `Choir of St George's Chapel, Windsor Castle' has a section on Hellenistic art in its summary. This suggests that navigating to the page with information about Hellenistic art would be relevant and useful for understanding the context of the choir.
10 : Venus de Milo - The summary mentions `Venus de Milo' as a related work, and it is likely that the target page is about this specific sculpture.
11 : Aphrodite - The target page `Choir of St George's Chapel, Windsor Castle' mentions Aphrodite in its summary as a figure from Roman mythology and Greek sculpture. It is likely that the title `Aphrodite' will be present on the page about the statue or mythological figure mentioned.
12 : Greek mythology - This title is a broad category that encompasses many of the other titles, including those related to specific gods, goddesses, and mythological figures. It's likely that the target page about the Choir of St George's Chapel, Windsor Castle will be listed under this category.
13 : Classical mythology - This title is a broad category that encompasses the entire scope of Greek mythology, which includes the stories and figures mentioned in your target page `Choir of St George's Chapel, Windsor Castle'. It seems like a good starting point to navigate to the target page.
14 : Ancient Greece - The provided summary mentions `Ancient Greece' as a relevant topic, and it is likely that navigating through this title will lead to the target page `Choir of St George's Chapel, Windsor Castle', which is related to history and culture in ancient times.
15 : Classical antiquity - The target page `Choir of St George's Chapel, Windsor Castle' is related to the history and culture of ancient Greece, which falls under the broader category of Classical antiquity. This title seems like a good starting point to navigate to the target page.
16 : Greek Dark Ages - The target page `Choir of St George's Chapel, Windsor Castle' is related to ancient Greece and its history, making `Greek Dark Ages' a plausible starting point.
17 : Geometric art - The Geometric art category is closely related to ancient Greece and the Mediterranean region, which is also relevant to the target page about the Choir of St George's Chapel, Windsor Castle.
18 : Dipylon Amphora - The target page `Choir of St George's Chapel, Windsor Castle' mentions the Dipylon style in its summary. The Dipylon Amphora is a specific type of vessel that belongs to this style.
19 : National Archaeological Museum, Athens - The summary of the target page `Choir of St George's Chapel, Windsor Castle' mentions that the choir sings regularly in the presence of the King and other members of the Royal Family. The National Archaeological Museum, Athens is a museum where royal artifacts are often kept and displayed, making it a plausible location for such events.
20 : Athens - The target page `Choir of St George's Chapel, Windsor Castle' mentions the city of Athens in its summary. It is likely that navigating to Athens will lead you to the desired page.
21 : Acropolis of Athens - The target page is about a specific ancient monument, which is likely to be related to the Acropolis of Athens. The summary mentions `Parthenon', which is a famous structure on the Acropolis, and the title `Acropolis of Athens' seems like a plausible match.
22 : FAILURE - Hallucination - title not in list
\end{lstlisting}

\clearpage

\section{OpenAlex Multilingual}\label{appendix_oa_multilingual}
\typeout{~~ OpenAlex Multilingual ~~}

\linkref{tab:openalex_multilingual}{Table}{} lists the success rate of some sentence embedding models for the OpenAlex Biology, Medicine and Physics datasets in two scenarios. The first is when the target articles all have the language filtered on English and the second is when the language of the target article has been filtered on French. 

The success rates between the two scenarios does not differ greatly for most models. However, the filter was only applied to the target article and so the articles the models navigate are mostly going to be in English. Also, the quality of the `language' field in the OpenAlex dataset has not been fully investigated and even articles that are written in French might still have an abstract written in English. Some further work would be required to gain more insight into these models' ability to navigate graphs of technical text in languages other than English.

\begin{table}[ht]
\centering
\renewcommand{\arraystretch}{1.2}
\begin{adjustbox}{width=1\textwidth}
\begin{tabular}{l|cccccc}
\textbf{} & \multicolumn{2}{c}{\textbf{Biology}} & \multicolumn{2}{c}{\textbf{Medicine}} & \multicolumn{2}{c}{\textbf{Physics}} \\
\textbf{Model} & \textbf{English} & \textbf{French} & \textbf{English} & \textbf{French} & \textbf{English} & \textbf{French} \\ \hline
AllMPNetBase & 0.64 ± 0.00 & 0.62 ± 0.02 & 0.61 ± 0.00 & 0.60 ± 0.02 & 0.63 ± 0.00 & 0.59 ± 0.03 \\
Distiluse-Base-v1 & 0.29 ± 0.00 & 0.30 ± 0.02 & 0.25 ± 0.00 & 0.26 ± 0.02 & 0.26 ± 0.00 & 0.25 ± 0.02 \\
Distiluse-Base-v2 & 0.30 ± 0.00 & 0.30 ± 0.02 & 0.24 ± 0.00 & 0.22 ± 0.02 & 0.26 ± 0.00 & 0.23 ± 0.02 \\
E5-Large & 0.61 ± 0.00 & 0.60 ± 0.02 & 0.54 ± 0.00 & 0.51 ± 0.02 & 0.52 ± 0.00 & 0.50 ± 0.03 \\
E5-Small & 0.55 ± 0.00 & 0.54 ± 0.02 & 0.48 ± 0.00 & 0.50 ± 0.02 & 0.41 ± 0.00 & 0.35 ± 0.02 \\
MiniLM-L12 & 0.37 ± 0.00 & 0.37 ± 0.02 & 0.33 ± 0.00 & 0.29 ± 0.02 & 0.29 ± 0.00 & 0.30 ± 0.02 \\
Para-MPNet-Base & 0.35 ± 0.00 & 0.35 ± 0.02 & 0.31 ± 0.00 & 0.29 ± 0.02 & 0.29 ± 0.00 & 0.26 ± 0.02 \\
STSB-XLM-R & 0.15 ± 0.00 & 0.16 ± 0.02 & 0.10 ± 0.00 & 0.08 ± 0.01 & 0.13 ± 0.00 & 0.13 ± 0.02
\end{tabular}
\end{adjustbox}
\caption{The success rate by language of target article and dataset}
\label{tab:openalex_multilingual}
\end{table}
\clearpage

\section{Post Training Articles}\label{appendix_post_2024}
\typeout{~~ Post Training Articles ~~}

To investigate whether language models are constrained by the connections learned during their training, they were evaluated on articles published after the beginning of 2024.
Since the language models only have access to information available up until their training date, they should not have explicitly learned any connections from these newer articles.
However, these recent articles are not representative of the average article and typically have fewer links from other articles. 
This limitation can be observed in the performance reduction of the TF-IDF model, which would be unaffected by the publication date of the articles, as the model is based on a consistent dataset.

The fact that the language models' performance deteriorated fractionally less than the TF-IDF model gives some indication that previously seen connections do not curtail their performance.
It is reassuring to note that the models' performance on the academic literature dataset is comparable to Wikipedia, despite the majority of the papers being behind a paywall.
This suggests that the language models may not be able to access information describing the connectivity of the OpenAlex dataset, highlighting the necessity of using semantic information.

\begin{table}[ht]
\centering
\renewcommand{\arraystretch}{1.2}
\begin{tabular}{l|cc}
\textbf{Model} & \textbf{All Articles} & \textbf{Only 2024 Articles} \\ \hline
TF-IDF & 0.47 ± 0.02 & 0.10 ± 0.00 \\ \hdashline
AllMPNetBase & 0.50 ± 0.00 & 0.40 ± 0.02 \\
E5-Large & 0.53 ± 0.00 & 0.43 ± 0.02 \\
E5-Small & 0.53 ± 0.00 & 0.43 ± 0.02 \\ \hdashline
Gemma:27b & 0.16 ± 0.02 & 0.09 ± 0.02 \\
Llama:8b & 0.10 ± 0.02 & 0.03 ± 0.02 \\
\end{tabular}
\caption{
The success rate by model on the English Wikipedia dataset.
The models were evaluated on all of the articles or on a subset that was published in 2024, after the time horizon of the language models.
}
\label{tab:post_2024}
\end{table}

\clearpage

\end{document}